\documentclass[11pt]{article}

\usepackage[utf8]{inputenc}   
\usepackage[T1]{fontenc}      
\usepackage{url}

\usepackage{helvet}           

\usepackage{amsmath}          
\usepackage{amssymb}          

\usepackage{graphicx}         
\usepackage{subcaption}       
\usepackage{xcolor}           
\usepackage{pgfplots}         
\pgfplotsset{compat=1.18}

\definecolor{myempirical}{RGB}{0, 128, 128} 
\definecolor{myoracle}{RGB}{60, 60, 60} 

\usepackage[round]{natbib}    

\usepackage[
    top=1in,
    bottom=1in, 
    left=1in,
    right=1in
]{geometry}

\usepackage{parskip}          
\usepackage{setspace}         
\usepackage{titlesec}

\titleformat{\subsubsection}[runin]
  {\normalfont\normalsize\bfseries}
  {\thesubsubsection}
  {0.5em}
  {}

\titlespacing*{\subsubsection}
  {0pt}      
  {1em}      
  {0.5em}    

\usepackage{enumitem}         
\setlist{
    topsep=0.2em,
    itemsep=0.2em,
    parsep=0.2em,
    leftmargin=1.5em
}

\usepackage{booktabs}         
\usepackage{longtable}        
\usepackage{array}            
\usepackage{ragged2e}         
\usepackage{threeparttable}

\newcolumntype{P}[1]{>{\RaggedRight}p{#1}} 

\usepackage{ulem}             

\usepackage{titling}
\title{Honesty in Causal Forests: When It Helps and When It Hurts}
\author{
    Yanfang Hou \quad Carlos Fern\'andez-Lor\'ia \\ 
    Hong Kong University of Science and Technology  
}
\date{}  

\begin{document}

\maketitle
\vspace{-3em}  



\begin{abstract}
Causal forests estimate how treatment effects vary across individuals, guiding personalized interventions in areas like marketing, operations, and public policy. A standard practice is \textit{honest estimation}---dividing the data into two samples, one to define subgroups and another to estimate treatment effects within them. This is intended to reduce overfitting and is the default in many software packages. But is it the right choice? We show that honest estimation can \textit{reduce} the accuracy of estimates of individual treatment effects---especially when effect heterogeneity is substantial and datasets are large enough to detect it. The reason is a bias-variance trade-off: honesty lowers the risk of overfitting but increases the risk of underfitting by limiting the data available to detect and model heterogeneity. Across more than 7,000 benchmark datasets, we find that the cost of using honesty by default can be as high as requiring 27\% more data to match the performance of models trained without it. Honesty is best understood as a form of regularization. Whether to adopt it should depend on the goals of the application and its empirical performance, not on reflexive default use.
\end{abstract}

\section{Introduction}

Some modeling choices become standard practice not because they are optimal, but because they are conservative, familiar, and built into the tools we use. But in applied settings, such defaults can come at the expense of making effective use of our data.

Causal forests provide a prominent example of this tension. They are a popular method for estimating how treatment effects vary across individuals \citep{wager2018estimation} and are increasingly used in fields such as marketing, operations, economics, and public policy to support personalization and targeting. Since 2019, dozens of papers in INFORMS journals have applied causal forests to experimental or observational data (see Appendix J), often as part of broader analytics pipelines that rely on individual-level effect estimation.

The method mirrors random forests for outcome prediction, but with causal effects as the target. Another key difference lies in how the data is used. Standard random forests use the full training sample to both build trees and estimate leaf-level predictions. Causal forests often divide the data in two: one part to form leaves, and another to estimate leaf-level effects. This practice---known as \textbf{honest estimation} or \textbf{honesty}---is the default in widely used software packages such as \texttt{grf} \citep{athey2019generalized} and \texttt{EconML} \citep{econml}, and is commonly recommended in applied work. But is this default a good idea?

On the one hand, honesty can help by preventing the same sampling error from influencing both leaf formation and effect estimation, reducing overfitting and selection bias~\citep{athey2016recursive}. On the other hand, it can hurt because it reduces the data available to form leaves and estimate effects.

We examine this trade-off in the context of estimating individual treatment effects as accurately as possible, a central concern in personalization and targeting applications. In this setting, honest estimation acts as a form of regularization: by constraining how much the model can adapt to the data, it trades flexibility for stability, much like other regularization choices. This trade-off can lead to less accurate models when effects differ substantially across individuals and the dataset is large enough to detect those differences.

To support this claim, we analyze how honesty affects the bias--variance trade-off in estimating individual treatment effects, both analytically and empirically. We find that its use of separate subsamples reduces variance, but the smaller subsample sizes limit the detection of heterogeneity and increase bias.

Our empirical analysis is based on 7,500 benchmark datasets from the Atlantic Causal Inference Conference (ACIC)~\citep{dorie2019automated}, covering a wide range of settings that vary in how difficult it is to model effect heterogeneity. The findings align with the mechanisms highlighted by our analytical results. Honesty works reasonably well when differences in effects are small or difficult to detect, but can backfire when heterogeneity is more pronounced. In such settings, honest estimation can require up to 27\% more data to achieve the same level of accuracy as causal forests learned without it.

This paper makes three contributions. First, we show that treating honest estimation as a default is unwarranted, as it can degrade accuracy when effect heterogeneity is substantial and detectable in the data. Second, we explain why this occurs by interpreting honesty as a form of regularization that governs a bias–variance trade-off. Third, we show that the choice between honest and non-honest estimation can matter substantially in practice, and offer concrete guidance on when to use each.

\section{How Tree-Based Causal Estimation Works}

This section defines the algorithm we consider for estimating treatment effects. We begin with a single tree, specifying how subpopulations are formed and how effects are estimated, and then extend to forests. Appendix I reviews the broader development of causal forests and honest estimation in the literature.

\subsection{Setup and Causal Quantities}

Consider the standard setup in which each individual is described by a feature vector $\mathbf{X}$ (e.g., age, income), a binary treatment assignment $T \in \{0, 1\}$, and an outcome $Y$. Each individual has two potential outcomes---$Y(1)$ if treated and $Y(0)$ if not---but we only observe one of them,
\begin{equation}
Y = T Y(1) + (1 - T) Y(0).
\label{eq:observed_outcome}
\end{equation}

We assume that treatment is assigned uniformly at random, so ignorability holds,
\begin{equation}
  Y(1), Y(0) \perp T \mid \mathbf{X}.  
\end{equation}
This means differences in outcomes between treated and control individuals can be interpreted causally. Our object of interest is the \textbf{individual treatment effect (ITE)},
\begin{equation}
    \text{ITE} = Y(1) - Y(0),
    \label{eq:ite}
\end{equation}
but effects can vary for reasons that are not predictable from $\mathbf{X}$. We therefore target the \textbf{conditional average treatment effect (CATE)},
\begin{equation}
\beta(\mathbf{x}) = \mathbb{E}[Y(1) - Y(0) \mid \mathbf{X} = \mathbf{x}],
\label{eq:cate}
\end{equation}
which captures the predictable component of effect variation. The CATE is the best any model can do, so we interpret it as the effect for any individual with feature values $\mathbf{x}$.

A tree estimates $\beta(\mathbf{x})$ by dividing the feature space into subpopulations (leaves) and estimating the average effect within each one. Let $\ell$ denote a leaf. The \textbf{subpopulation average treatment effect (SPATE)} of $\ell$ is:
\begin{equation}
    \tau(\ell) = \mathbb{E}[Y(1) - Y(0) \mid \mathbf{X} \in \ell].
    \label{eq:leaf_cate}
\end{equation}
Given an \textbf{estimation sample} $\mathcal{S}_{\text{es}}$, the tree estimates the SPATE by comparing the average outcomes of treated and control units:
\begin{equation}
    \hat\tau(\ell; \mathcal{S}_{\text{es}}) =
    \frac{1}{n_\ell^1} \sum_{\substack{i \in \mathcal{S}_{\text{es}}, \\ \mathbf{x}_i \in \ell,\, t_i = 1}} y_i
    ~~-~~
    \frac{1}{n_\ell^0} \sum_{\substack{i \in \mathcal{S}_{\text{es}}, \\ \mathbf{x}_i \in \ell,\, t_i = 0}} y_i,
    \label{eq:leaf_estimate}
\end{equation}
where $n_\ell^1$ and $n_\ell^0$ are the number of treated and control units in the leaf.

\subsection{How Subpopulations Are Chosen}

We want subpopulations such that the SPATE $\tau(\ell)$ closely approximates the CATEs $\beta(\mathbf{x})$ of individuals with $\mathbf{x} \in \ell$. A subpopulation is therefore ``good'' if effects vary little within it. For any candidate leaf $\ell$, we measure this using the approximation's mean squared error (MSE),
\begin{equation}
\text{MSE}(\ell) = \mathbb{E} \left[ (\beta(\mathbf{X}) - \tau(\ell))^2 \mid \mathbf{X} \in \ell \right].
\label{eq:node_mse}
\end{equation}

Learning the tree then amounts to recursively finding splits that reduce this within-leaf error. For a split of a parent node $\ell_0$ into children $\ell_1$ and $\ell_2$, the MSE gain is
\begin{equation}
\begin{aligned}
\text{Gain} 
&= \text{MSE}(\ell_0) \\
&\quad - \sum_{i=1}^{2} \mathbb{P}(\mathbf{X} \in \ell_i \mid \mathbf{X} \in \ell_0) \cdot \text{MSE}(\ell_i).
\end{aligned}
\label{eq:mse_gain}
\end{equation}
The tree learner uses a \textbf{splitting sample} $\mathcal{S}_{\text{sp}}$ and the following criterion to estimate this gain:
\begin{equation}
\text{Criterion} =
\frac{n_{\ell_1} n_{\ell_2}}{(n_{\ell_1} + n_{\ell_2})^2}
\left( \hat\tau(\ell_1; \mathcal{S}_{\text{sp}}) - \hat\tau(\ell_2; \mathcal{S}_{\text{sp}}) \right)^2,
\label{eq:split_criterion}
\end{equation}
where $n_{\ell_i}$ is the number of observations in leaf $\ell_i$ in the splitting sample $\mathcal{S}_{\text{sp}}$. This criterion favors splits that form groups with different estimated effects while penalizing highly unbalanced splits. It is a consistent estimate of the MSE gain (proof in Appendix A). 

We denote by $\hat\ell(\mathbf{x}; \mathcal{S}_{\text{sp}})$ the \textbf{leaf assignment} of an individual $\mathbf{x}$ under the tree partition learned from $\mathcal{S}_{\text{sp}}$. 

\subsection{Effect Estimation}

Given a splitting sample $\mathcal{S}_{\text{sp}}$ and an estimation sample $\mathcal{S}_{\text{es}}$, the tree assigns an individual $\mathbf{x}$ to a leaf $\hat\ell(\mathbf{x}; \mathcal{S}_{\text{sp}})$ and uses the corresponding SPATE estimate as the individual’s effect:
\begin{equation}
\hat\beta^{\text{tree}}(\mathbf{x}; \mathcal{S}_{\text{sp}}, \mathcal{S}_{\text{es}})
=
\hat\tau\!\left(\hat\ell(\mathbf{x}; \mathcal{S}_{\text{sp}}); \mathcal{S}_{\text{es}}\right).
\label{eq:tree_estimator}
\end{equation}

Under \textbf{honest estimation~(HE)}, the data is divided so that $\mathcal{S}_{\text{sp}}$ and $\mathcal{S}_{\text{es}}$ are disjoint subsamples. Under \textbf{adaptive estimation (AE)}, all the data is used for both leaf formation and effect estimation, so 
$\mathcal{S}_{\text{sp}}=\mathcal{S}_{\text{es}}$.

A causal forest extends this by aggregating estimates across many trees, each constructed from a different subsample of the training data. Let $\{\mathcal{S}^{(b)}\}_{b=1}^B$ denote $B$ subsamples drawn without replacement. The prediction of the tree trained on subsample $\mathcal{S}^{(b)}$ is
\begin{equation}
    \hat\beta_b^{\text{tree}}(\mathbf{x}) = \hat\beta^{\text{tree}}(\mathbf{x}; \mathcal{S}^{(b)}_{\text{sp}}, \mathcal{S}^{(b)}_{\text{es}}). 
\end{equation}
The forest prediction $\hat\beta(\mathbf{x})$ is the average of these tree predictions:
\begin{equation}
\hat\beta(\mathbf{x})
=
\frac{1}{B}
\sum_{b=1}^{B}
\hat\beta_b^{\text{tree}}(\mathbf{x}).   
\end{equation}

\section{A Bias–Variance Perspective}

To unpack the trade-off between HE and AE, we analyze how each affects the bias and variance of causal forest estimates. For a given individual $\mathbf{x}$ with CATE $\beta(\mathbf{x})$, the MSE decomposes as
\begin{equation}
\text{MSE}(\hat\beta) = \left( \mathbb{E}[\hat\beta] - \beta \right)^2 + \text{Var}\left(\hat\beta\right).
\end{equation}

Our goal is not to derive precise conditions under which HE dominates AE, or vice versa. The choice between the two should be guided empirically, as discussed in Section~\ref{sec:empirical}. Instead, we use the bias--variance decomposition to clarify \emph{how} honesty affects the accuracy of CATE estimates.

Because the bias and variance of a causal forest arise from aggregating individual trees, we begin by understanding the errors made by a single tree:
\begin{equation}
\begin{aligned}
&\hat\beta^{\text{tree}}(\mathbf{x},\mathcal{S}_{\text{sp}}, \mathcal{S}_{\text{es}})\\
&=\beta
+ \underbrace{\tau(\hat\ell)-\beta}_{\text{approximation error}}
+ \underbrace{\hat\tau(\hat\ell)-\tau(\hat\ell)}_{\text{estimation error}},
\end{aligned}
\label{eq:approx-est}
\end{equation}
where $\hat\ell$ and $\hat\tau(\cdot)$ are shorthand for 
$\hat\ell(\mathbf{x};\mathcal{S}_{\text{sp}})$ and 
$\hat\tau(\cdot;\mathcal{S}_{\text{es}})$. \textbf{Approximation error} comes from using a SPATE to approximate individual CATEs.  For example, if a leaf groups together individuals aged 20--40, the SPATE may poorly represent the CATE of a 38-year-old whose response differs from others in the same leaf. \textbf{Estimation error} reflects how well we estimate the SPATE. Even within a leaf with homogeneous CATEs, the sample may not be representative of the subpopulation---due to sampling variability or selection bias---leading to inaccurate effect estimates. We analyze next how these errors contribute to bias and variance.

\subsection{Where Bias Comes From}

The bias of a causal forest coincides with the bias of its individual trees. Taking expectations,
\begin{equation}
\mathbb{E}\!\left[\hat\beta\right]
=
\frac{1}{B}\sum_{b=1}^B \mathbb{E}\!\left[\hat\beta_b^{\text{tree}}\right]
=
\mathbb{E}\!\left[\hat\beta_b^{\text{tree}}\right],
\end{equation}
because the trees are identically distributed. Forest bias is the same as tree bias, so we focus on tree bias.

Using the decomposition in Equation~\eqref{eq:approx-est}, we can write tree bias as the sum of its expected approximation and estimation errors:
\begin{equation}\label{eq:error_decomposition}
\begin{aligned}
&\mathbb{E}[\hat\beta^{\text{tree}}(\mathbf{x}; \mathcal{S}_{\text{sp}}, \mathcal{S}_{\text{es}})] - \beta \\
&= \underbrace{\mathbb{E}_{\hat\ell}\left[\tau\left(\hat\ell\right)\right] - \beta}_{\text{Approximation bias}} 
 + \underbrace{
\mathbb{E}_{\hat\ell} \left[ \mathbb{E}_{\hat\tau} \left[
\hat\tau\left(\hat\ell\right) \mid \hat\ell 
\right] - \tau\left(\hat\ell\right) \right]
}_{\text{Estimation bias}}.
\end{aligned}
\end{equation}

Estimation bias arises under AE because the same data is used to select splits and estimate effects. The tree learner searches over many candidate splits and chooses those that appear to best separate individuals by treatment effects, even when that separation reflects sampling error rather than true heterogeneity. As a result, SPATE estimates under AE are systematically exaggerated---an instance of selection bias---with above-average effects biased upward and below-average effects biased downward.

HE eliminates estimation bias by construction. By using separate samples for splitting and estimation, leaf assignments and effect estimates become independent. Even if some splits are chosen due to sampling error, the effect estimate within each leaf remains unbiased:
\begin{equation}
\mathbb{E} \left[ \hat\tau\left(\hat\ell\right) \mid \hat\ell \right]
= \tau\left(\hat\ell\right).
\label{eq:honest_unbiased}
\end{equation}

However, data splitting under HE increases approximation bias. With less data to evaluate candidate splits, the tree learner has a harder time identifying informative features, so the resulting subpopulations may align less closely with heterogeneity in treatment effects.

The implication is that HE and AE trade off approximation and estimation bias in opposite ways. Empirically, we find that AE wins in this trade-off. We illustrate this next with a simple analytical example.

\subsection{A Stylized Example of Bias}\label{sec:example}

Consider a deliberately simple setting---a tree with a single split---designed to cleanly isolate the sources of bias. Formal derivations are in Appendix B.

Suppose $\mathbf{X}$ is a vector of binary features. Exactly one feature, $X_1$, is informative about treatment effects:
\begin{equation}
\beta(\mathbf{x}) = \theta (2x_1 - 1), \quad \theta > 0.
\end{equation}
Individuals with $x_1 = 1$ benefit from treatment, while those with $x_1=0$ are harmed by the same amount. All remaining features serve only as noise. The average treatment effect in the population is zero.

Individual effects are estimated using a tree with a single split. If the tree splits on the feature $X_1$, the resulting leaves align perfectly with true effect heterogeneity. If it splits on any other feature, both leaves have the same SPATE---equal to zero---and are therefore uninformative. 

Consider an individual with $x_1=1$, for whom the CATE is $\theta$ (the case $x_1=0$ is symmetric). The superscripts H and A distinguish HE and AE quantities, and $\hat\ell=1$ denotes splitting on $X_1$. As discussed earlier, HE has no estimation bias, so leaf‑level (SPATE) estimates are unbiased:
\begin{equation}
\mathbb{E}\left[\hat{\beta}^{\text{H}} \mid \hat\ell^\text{~H} \right] = \tau(\hat\ell^\text{~H}).
\end{equation}

However, HE is still biased for CATE estimation because it may split on a non-informative feature. Taking expectations over splits,
\begin{equation*}
\mathbb{E}\left[\hat\beta^{\text{H}}\right]
= \mathbb{P}(\hat\ell^\text{~H} = 1) \cdot \theta + \mathbb{P}(\hat\ell^\text{~H} \neq 1) \cdot 0.
\end{equation*}
The resulting bias is:
\begin{equation}
\mathbb{E}[\hat\beta^{\text{H}}] - \beta
= \theta \cdot \left( \mathbb{P}(\hat\ell^\text{~H} = 1) - 1 \right),
\end{equation}
which is negative unless HE selects the informative feature with probability one. This corresponds entirely to approximation bias.

Now consider AE. When the tree splits on a non‑informative feature ($\hat\ell^\text{A} \ne 1$), the actual SPATE in both leaves is zero. AE systematically exaggerates the SPATE estimates, but the errors are symmetric around zero and cancel out in expectation, so there is no estimation bias:
\begin{equation}\label{eq:adaptive_bias_uninformative}
\mathbb{E}\left[\hat{\beta}^{\text{A}} \mid \hat\ell^\text{A} \ne 1 \right] = 0.
\end{equation}

Things play out differently when AE splits on the informative feature ($\hat\ell^\text{A} = 1$). Sampling error can make this split look more or less informative than it truly is, and the tree learner is more likely to select it when the error is favorable. Conditioning on $\hat\ell^\text{A} = 1$ therefore selects cases where the estimation error is positive or not too negative---this is the selection bias discussed earlier. Because AE uses the same data for splitting and estimation, this bias carries over to the leaf estimates, leading to upward bias:\footnote{The bias is upward for individuals with $x_1=1$ and downward for those with $x_1=0$.}
\begin{equation}\label{eq:adaptive_bias_informative}
\mathbb{E}\left[\hat{\beta}^{\text{A}} \mid \hat\ell^\text{A} = 1 \right] \geq \theta.
\end{equation}

AE's advantage is that it splits on the informative feature more often than HE,
\begin{equation}\label{eq:adaptive_probability}
\mathbb{P}(\hat\ell^\text{A} = 1) \ge \mathbb{P}(\hat\ell^\text{~H} = 1).
\end{equation}
This means AE accepts some estimation bias in exchange for lower approximation bias. Its overall bias is negative and always smaller than that of HE in this example:
\begin{equation}\label{eq:adaptive_lower_bias}
\left| \mathbb{E}[\hat{\beta}^{\text{A}}] - \beta \right|
\leq \left| \mathbb{E}[\hat{\beta}^{\text{H}}] - \beta \right|.
\end{equation}
Figure~\ref{fig:bias-illustration} illustrates this trade‑off.

\begin{figure}
\centering
\begin{tikzpicture}
\begin{axis}[
    width=12cm,
    height=8cm,
    ymin=0, ymax=4.5,
    xmin=0.5, xmax=2.5,
    xtick={1,2},
    xticklabels={Honest Estimation (HE), Adaptive Estimation (AE)},
    ytick={0, 1, 2, 3, 4},
    axis x line=bottom,
    axis y line=left,
    ylabel={Expected effect estimate},
    legend pos=south east,
    legend style={font=\small},
    ymajorgrids=true,
    grid style=dashed,
    clip=false
]

\addplot[
    only marks,
    mark=*,
    mark size=2.5pt,
    color=blue
] coordinates {
    (1, 3.5) 
    (2, 4.1) 
};
\addlegendentry{Good split ($\hat\ell=1$)}

\addplot[
    only marks,
    mark=*,
    mark size=2.5pt,
    color=black
] coordinates {
    (1, 0) 
    (2, 0) 
};
\addlegendentry{Bad split ($\hat\ell\neq1$)}

\addplot[
    only marks,
    mark=*,
    mark size=2.5pt,
    color=red
] coordinates {
    (1, 2)   
    (2, 2.7) 
};
\addlegendentry{Average over splits}

\addplot[dashed, gray, thick, domain=0.5:2.5] {3.5};
\addlegendentry{CATE}

\addplot[
    color=orange!80!black,
    ultra thick,
    dotted
] coordinates {
    (1, 2) (1, 3.5)
};
\addplot[
    color=orange!80!black,
    ultra thick,
    dotted
] coordinates {
    (2, 2.7) (2, 3.5)
};
\addlegendentry{Bias}

\end{axis}
\end{tikzpicture}
\caption{\textbf{Bias comparison of HE and AE.} HE is unbiased conditional on a correct split, while AE exhibits selection bias. However, AE selects the informative feature more often, reducing approximation bias and producing an estimate closer to the CATE in expectation.}
\label{fig:bias-illustration}
\end{figure}

\subsection{Where Variance Comes From}\label{sec:variance}

We now turn to the variance of a causal forest. Because a forest averages predictions across trees, its variance corresponds to the average of tree covariances:
\begin{equation}
\mathrm{Var}\!\big(\hat\beta(x)\big)
=
\frac{1}{B^2}
\sum_{b=1}^B
\sum_{b'=1}^B
\mathrm{Cov}\!\big(\hat\beta_b^{\text{tree}},\hat\beta_{b'}^{\text{tree}}\big).
\label{eq:forest-variance}
\end{equation}
As the number of trees grows, the dominant contribution comes from cross-tree covariances, $\mathrm{Cov}(\hat\beta_b,\hat\beta_{b'})$ for $b\neq b'$. Understanding forest variance therefore reduces to understanding the sources of dependence between tree predictions.

We first decompose each tree prediction into the two components introduced in Equation~\eqref{eq:approx-est}: an approximation target and an estimation error.
Tree $b$ assigns $\mathbf{x}$ to a leaf $\hat\ell_b = \hat\ell(\mathbf{x}; \mathcal{S}^{(b)}_{\text{sp}})$, which defines a SPATE
\begin{equation}
\tau_b
=
\tau\!\left(\hat\ell_b \right).
\label{eq:tree-spate}
\end{equation}
This is the \textbf{approximation target}. The corresponding prediction is
\begin{equation}
\hat\beta_b^{\text{tree}}
=
\tau_b
+
\varepsilon_b,
\qquad
\varepsilon_b
=
\hat\tau\!\left(\hat\ell_b;\mathcal{S}^{(b)}_{\text{es}}\right)
-
\tau_b,
\label{eq:tree-variance-decomp}
\end{equation}
where $\varepsilon_b$ is
the \textbf{estimation error}.

We then decompose the covariance between tree predictions as
\begin{equation}
\begin{aligned}
\mathrm{Cov}\!\big(\hat\beta_b^{\text{tree}},\hat\beta_{b'}^{\text{tree}}\big)
&=
\underbrace{\mathrm{Cov}\!\big(\tau_b,\tau_{b'}\big)}_{\text{Target coupling}}
+
\underbrace{\mathrm{Cov}\!\big(\varepsilon_b,\varepsilon_{b'}\big)}_{\text{Noise overlap}}\\
&\quad +
\underbrace{\mathrm{Cov}\!\big(\tau_b,\varepsilon_{b'}\big)
+
\mathrm{Cov}\!\big(\varepsilon_b,\tau_{b'}\big)}_{\text{Target--estimation spillover}}.
\label{eq:covariance-decomposition}
\end{aligned}
\end{equation}
This decomposition reveals three distinct sources of dependence. We discuss each in turn and provide analytical derivations to support our arguments in Appendix C. 

\subsubsection{Target Coupling,}$\mathrm{Cov}\!\big(\tau_b,\tau_{b'}\big)$, captures dependence across trees arising from variability in their approximation targets. To build intuition, consider the stylized example from Section~\ref{sec:example}. A tree may sometimes split on the informative feature, producing leaves that reflect true effect heterogeneity, and sometimes fail to do so, producing uninformative leaves. Which occurs depends on the specific training sample. Because trees are grown on overlapping subsamples, they tend to succeed or fail together. This shared fate induces covariance between their approximation targets.

HE affects target coupling indirectly by reducing the size of the splitting sample, limiting trees’ ability to identify good approximation targets. Whether this increases or decreases target coupling depends on the \textbf{signal-to-noise ratio (SNR)}: how easy it is to infer variation in treatment effects from data.

When informative features are easy to identify (high SNR), HE disrupts otherwise reliable splits, making approximation targets more sensitive to the training sample and increasing target coupling. When informative features are difficult to detect reliably (low SNR), trees find good approximation targets only inconsistently. HE stabilizes them by removing those unreliable attempts, reducing target coupling---though at the cost of also missing the occasional good split. Figure~\ref{fig:target-coupling} illustrates this effect in our stylized example.

Empirically, however, we find that HE produces lower target coupling than AE even when SNR is high.

\begin{figure}[t]
\centering
\begin{tikzpicture}
\begin{axis}[
    width=0.7\textwidth,
    height=0.45\textwidth,
    xlabel={Probability of Selecting an Informative Split},
    ylabel={Strength of Target Coupling},
    xmin=0, xmax=1.05,
    ymin=0, ymax=0.26,
    domain=0:1,
    samples=200,
    axis lines=left,
    ticks=none,
    clip=false
]

\addplot[thick] {x*(1-x)};

\addplot[dashed] coordinates {(0.38,0) (0.38,0.045)};
\addplot[dashed] coordinates {(0.18,0) (0.18,0.045)};

\draw[->, thick]
(axis cs:0.38,0.05) -- (axis cs:0.18,0.05)
node[midway, above, text width=4cm, align=center] {\scriptsize HE decreases target coupling};

\node[anchor=north] at (axis cs:0.12,0.2) {Low SNR};

\addplot[dashed] coordinates {(0.83,0) (0.83,0.045)};
\addplot[dashed] coordinates {(0.63,0) (0.63,0.045)};

\draw[->, thick]
(axis cs:0.83,0.05) -- (axis cs:0.63,0.05)
node[midway, above, text width=4cm, align=center] {\scriptsize HE increases target coupling};

\node[anchor=north] at (axis cs:0.9,0.2) {High SNR};

\addplot[only marks, mark=*, mark size=2pt] coordinates {(0.18,0)};
\addplot[only marks, mark=*, mark size=2pt] coordinates {(0.38,0)};
\addplot[only marks, mark=*, mark size=2pt] coordinates {(0.63,0)};
\addplot[only marks, mark=*, mark size=2pt] coordinates {(0.83,0)};

\end{axis}
\end{tikzpicture}
\caption{
\textbf{Target coupling as a function of the SNR.}
HE decreases the probability of an informative split compared to AE. In high-SNR settings (right), this moves trees away from reliable and informative splits, increasing sampling sensitivity and target coupling. In low-SNR settings (left), the same shift moves trees toward stable, uninformative targets and reduces target coupling.}
\label{fig:target-coupling}
\end{figure}

\subsubsection{Target--Estimation Dependence}is the source of variance that honesty is designed to address. This dependence arises when the same sampling error that favors a split also biases the estimated SPATE in the same direction: above-average SPATEs are more likely to be overestimated, while below-average SPATEs are more likely to be underestimated. This introduces a dependence between the approximation target and the estimation error that inflates variance.

By separating the samples used for splitting and effect estimation, HE breaks this dependence within a tree, so the estimation error has conditional mean zero given the leaf assignment:
\begin{equation}
\mathbb{E}[\varepsilon_b \mid \hat\ell_b] = 0.
\label{eq:honesty-within-tree}
\end{equation}
This rules out target--estimation dependence \emph{within} a tree, but not at the forest level. Trees are trained on overlapping subsamples of the same data, so observations in one tree's estimation sample may appear in another's splitting sample. Therefore, the estimation error of one tree may correlate with the approximation target of another, resulting in the \textbf{target--estimation spillover} in Equation~\eqref{eq:covariance-decomposition}.

To eliminate this dependence entirely, the approximation target of each tree must be independent of the estimation error across \emph{all} trees.\footnote{One way to achieve this is to partition the original training data once into disjoint splitting and estimation subsets, and then subsample separately from each subset when learning every tree. However, we do not recommend this design because it would likely amplify target coupling and noise overlap.} Under full independence,
\begin{equation}
\mathbb{E}[\varepsilon_b \mid \hat\ell_b, \hat\ell_{b'}] = 0.
\label{eq:idealized-honesty}
\end{equation}

We refer to this idealized condition as \textbf{forest honesty}. While honest causal forests do not satisfy forest honesty exactly, they are designed to approximate it by separating target selection from effect estimation, and thus more closely adhere to this condition than adaptive forests. Assumptions of this kind are common in the theoretical analysis of tree-based methods, where they are introduced to simplify dependence and make analysis tractable. For example, \cite{scornet2016asymptotics} study forests with data-independent (uniform) splitting, and \cite{klusowski2021sharp} use midpoint splitting rules.

Forest honesty plays the same role here. We next analyze the covariance decomposition in Equation~\eqref{eq:covariance-decomposition} under this abstraction to isolate the effect of reducing target--estimation dependence.

\subsubsection{How Honesty Reduces Overfitting.}

Under forest honesty, the \textbf{target--estimation spillover} terms in Equation~\eqref{eq:covariance-decomposition} disappear:
\begin{equation}
\mathrm{Cov}\!\big(\tau_b,\varepsilon_{b'}\big)
=
\mathrm{Cov}\!\big(\varepsilon_b,\tau_{b'}\big)
=
0.
\label{eq:no-target-estimation-dependence}
\end{equation}
 
This leaves \textbf{noise overlap},
$\mathrm{Cov}(\varepsilon_b,\varepsilon_{b'})$, which arises because different trees may reuse the same observations in their estimation
samples, introducing correlation in their estimation errors. It has two components:
\begin{equation}
\begin{aligned}
&\mathrm{Cov}\big(\varepsilon_b,\varepsilon_{b'}\big)\\
&=
\mathbb{E}\Big[
\mathrm{Cov}\big(\varepsilon_b,\varepsilon_{b'} \mid \hat\ell_b, \hat\ell_{b'}\big)
\Big]\\
&\quad +
\mathrm{Cov}\Big(
\mathbb{E}\big[\varepsilon_b \mid \hat\ell_b, \hat\ell_{b'}\big],
\mathbb{E}\big[\varepsilon_{b'} \mid \hat\ell_b, \hat\ell_{b'}\big]
\Big).
\end{aligned}
\label{eq:noise-overlap-totcov}
\end{equation}

The first is unavoidable and depends primarily on sample size and how similar the leaves are across trees. Smaller samples and more similar leaves imply greater overlap. Because forests average over many trees, the reduced estimation sample size induced by HE does not affect this component asymptotically.

The second arises from target--estimation dependence. The same sampling error that makes a partition look informative in one tree is likely to make it look informative in other trees too. When those noisy observations are reused for estimation, as under AE, it induces additional correlation between estimation errors. Under forest honesty, this component vanishes, reducing variance.

\subsection{Summary}

Estimating individual treatment effects with trees involves two sources of error: approximation error, from summarizing effects within leaves, and estimation error, from sampling variability and selection bias in estimating those leaf-level effects. HE and AE trade off these errors differently.

HE eliminates estimation bias by separating the data used for splitting and estimation, but increases approximation bias by limiting trees' ability to detect effect heterogeneity. AE accepts some estimation bias in exchange for lower approximation bias, which leads to lower overall bias. 

At the forest level, variance arises from dependence across trees. HE reduces variance by breaking the link between sampling error in split selection and effect estimation. In high-SNR settings, it can theoretically increase variance by destabilizing otherwise reliable splits, but we do not observe this empirically.

Overall, honesty acts as a form of regularization: more bias, less variance. Theory predicts it helps when SNR is low and hurts when it is high.

\section{Empirical Study}\label{sec:empirical}

This section addresses three practical questions. First, how should a practitioner choose between HE and AE? Second, how much does that choice matter? Third, do their performance patterns match the bias--variance mechanisms developed earlier?

To mirror applied settings where treatment effects are unobserved, we adopt a fully data-driven workflow. We tune HE and AE via cross-validation using transformed outcomes~\citep{athey2015machine} as effect proxies, and then evaluate performance using ground-truth CATEs. We also quantify what the choice costs by asking how much additional data one approach requires to match the other.

\subsection{Benchmark Data}

We evaluate HE and AE using the 2016 Atlantic Causal Inference Conference (ACIC) benchmarking suite~\citep{dorie2019automated}, which was created for a competition to compare causal estimators. The benchmark consists of semi-synthetic datasets that combine real covariates from the Infant Health and Development Program (IHDP) with simulated treatment assignments and potential outcomes, providing ground-truth CATEs for each individual. 

The suite includes 77 distinct types of data-generating processes (DGPs). Within each type, 100 datasets are generated by varying the underlying parameters. We exclude datasets with no treatment effect heterogeneity, yielding 7,500 datasets for analysis. Each dataset contains 4,802 observations and 58 covariates. The DGPs vary widely in functional form, complexity, and the strength of treatment effect heterogeneity, so the benchmark is well suited to study the HE--AE trade-off in a multitude of conditions. The datasets were originally designed to study causal estimation under confounding, so we reassign treatment uniformly at random to isolate the task of modeling effect heterogeneity under clean randomization.

We calculate the SNR for each dataset as follows. We define noise as idiosyncratic variation in outcomes,
\begin{equation}
    \xi = Y - \mathbb{E}[Y\mid \mathbf{X}, T].
\end{equation}
The SNR for a given dataset is
\begin{equation}\label{eq:SNR}
\text{SNR} = \frac{\mathrm{Var}(\beta(\mathbf{X}))}{\mathrm{Var}(\xi)}.    
\end{equation}
Higher SNR implies treatment effect heterogeneity is stronger and easier to estimate from data.

\subsection{Design and Workflow}\label{sec:design}

Our empirical design mirrors applied deployment: treatment effects are unobserved, so model selection must rely on observable quantities. For each dataset, we randomly assign treatment with probability 0.5 and split the sample into 4,000 observations for training and 802 for testing. 

\paragraph{Tuning.}
We tune AE and HE separately using 5-fold cross-validation on the training sample.\footnote{We use the implementation from \texttt{EconML} \citep{econml}. Each forest consists of 100 trees trained on subsamples of half the training data. We set other hyperparameters that affect model complexity, such as maximum depth, to non-binding values so that regularization is governed by the \texttt{min\_samples\_leaf} parameter.} The tuning parameter is \texttt{min\_samples\_leaf}. Because HE uses only half of each subsample for splitting, it requires smaller minimum leaf sizes to attain a comparable number of leaves to AE. The tuning grid for AE is \(\{10,20,40,80,120,160,240,320\}\) and
\(\{5,10,20,40,60,80,120,160\}\) for HE (half the AE values).

Because individual CATEs ($\beta_i$) are unobserved, the tuning relies on the transformed outcome $z_i$, whose conditional expectation equals the CATE~\citep{athey2015machine}. We select hyperparameters by minimizing the MSE between the predicted CATE ($\hat\beta_i$) and $z_i$ on the validation set $\mathcal{S}_{\text{val}}$:
\begin{equation}
\begin{aligned}
\text{MSE}_{\text{cv}}&=\frac{1}{|\mathcal{S}_{\text{val}}|}\sum_{i\in\mathcal{S}_{\text{val}}}(\hat\beta_i-z_i)^2, \\
z_i&=\frac{t_iy_i}{\omega(\mathbf{x}_i)}-\frac{(1-t_i)y_i}{1-\omega(\mathbf{x}_i)}
\end{aligned}
\label{eq:transformed_mse}
\end{equation}
where $\omega(\mathbf{x})=\Pr(T=1\mid \mathbf{X}=\mathbf{x})$. In our experimental design, $\omega(\mathbf{x}_i)=0.5$. 

\paragraph{Selection.}
The comparison between AE and HE only matters when the methods can detect heterogeneity in the first place. Our selection procedure has two steps. First, we create a new 5-fold split of the training data---distinct from the one used for tuning---and apply the heterogeneity test of \cite{imai2025statistical}, in its cross-fitting implementation, to both tuned estimators.\footnote{Appendix D.5 details the implementation of the test. As an empirical check, its false positive rate on the 200 ACIC datasets with no true heterogeneity is 1.5\% for both AE and HE, below the nominal 5\% level.} If neither passes, it means the signal is too weak to support reliable effect estimation, so we exclude the dataset. If only one passes, we select it. If both pass, we compare their performance using out-of-fold transformed-outcome MSE: we treat HE as the default and select AE only if its MSE is lower by at least one standard error, estimated from the paired differences in squared errors across held-out units.

The asymmetry of the rule is theoretically motivated. HE is expected to outperform AE in low-SNR settings, where the empirical comparison is also least reliable, so we default to HE and let the data overturn that default only when the evidence is strong. The one-standard-error threshold follows a standard selection heuristic~\citep{breiman1984classification, hastie2009elements}, but it is a conventional choice; other thresholds could be used. Appendix D develops the full rationale and shows that performance is robust across thresholds.

\paragraph{Evaluation.} We evaluate performance using a metric we denote \( S^2 \), defined in analogy to \( R^2 \). While \( R^2 \) measures the proportion of outcome variance explained by a predictive model, \( S^2 \) measures the proportion of explainable treatment effect variance---the signal---captured by a causal model. It is based on the MSE between the predicted and true CATEs:
\begin{equation}
\text{MSE}_{\text{te}} = \frac{1}{|\mathcal{S}_{\text{te}}|} \sum_{i \in \mathcal{S}_{\text{te}}} (\hat{\beta}_i - \beta_i)^2, 
\end{equation}
where $\mathcal{S}_{\text{te}}$ denotes a test set. We then define:
\begin{equation}
S^2 = 1 - \frac{\text{MSE}_{\text{te}}}{\text{Var}_{\text{te}}(\beta)},
\end{equation}
where $\mathrm{Var}_{\text{te}}(\beta)$ is the variance of the CATEs in the test set. This normalization makes results comparable across datasets with different levels of heterogeneity; higher $S^2$ implies the model captures a larger share of the explainable variation in treatment effects.

\subsection{Results}

Among the 7,500 ACIC datasets, neither tuned forest passes the heterogeneity test in 496 cases. We therefore evaluate AE, HE, and the empirical selector on the remaining 7,004 datasets.

We evaluate our selection strategy using a regret framework. For a given dataset, regret measures the $S^2$ loss from using a strategy $m$ instead of the oracle (the best of AE and HE in hindsight): 
\begin{equation}
\text{Regret}(m)=\max(S^2_{\text{AE}},S^2_{\text{HE}})-S^2_m.
\end{equation}

Average regret decomposes into how often a strategy picks wrong and how costly those mistakes are:
\begin{equation}
\mathbb{E}[\text{Regret}]
=
\text{Error rate}
\times
\mathbb{E}[\text{Regret}\mid \text{error}].
\end{equation}

\begin{table}[t]
\centering
\caption{Selection performance across 7,004 ACIC datasets.}
\label{tab:selection_performance}
\begin{threeparttable}
\setlength{\tabcolsep}{13pt}
\begin{tabular}{lccc}
\toprule
Method 
& $\mathbb{E}[\text{Regret}]$ 
& Error rate 
& $\mathbb{E}[\text{Regret}\mid\text{error}]$ \\
\midrule
Empirical & \textbf{0.0124} & 30.71\% & \textbf{0.0405} \\
AE & 0.0136 $(+0.0012^{**})$ & \textbf{25.20\%} & 0.0542 $(+0.0137^{***})$ \\
HE & 0.0344 $(+0.0220^{***})$ & 74.80\% & 0.0460 $(+0.0055^{***})$ \\
\bottomrule
\end{tabular}
\begin{tablenotes}[flushleft]
\footnotesize
\item Notes: Values in parentheses report difference relative to the empirical selector; positive values indicate higher regret. $^{**}p<0.05$, $^{***}p<0.01$.
\item Differences in $\mathbb{E}[\text{Regret}] $ are tested using one-sided paired $t$-tests across datasets. Differences in $\mathbb{E}[\text{Regret}\mid\text{error}] $ are tested with bootstrapping because the conditioning events differ across methods.
\end{tablenotes}
\end{threeparttable}
\end{table}

Table~\ref{tab:selection_performance} summarizes performance across the 7,004 datasets. AE has lower average regret than HE and is more often the better choice, but its mistakes carry more downside risk. The empirical selector achieves the lowest average regret and reduces the cost of mistakes, but has a higher error rate than AE. Both improvements are statistically significant relative to AE and HE.\footnote{As a robustness check, we repeat the full procedure under five independent treatment randomizations. The empirical selector achieves lower $\mathbb{E}[\text{Regret}] $ and $\mathbb{E}[\text{Regret}\mid\text{Error}]$ than both AE and HE in every run. The gains are significant for $\mathbb{E}[\text{Regret}] $ against HE in all five runs and against AE in four, and for $\mathbb{E}[\text{Regret}\mid\text{Error}]$ against AE in all five runs and against HE in four.}

\begin{figure}[t]
    \centering
    \begin{tikzpicture}
    \begin{axis}[
        width=12cm, 
        height=8cm,
        xlabel={Percentile},
        ylabel={Regret},
        ymin=0., ymax=0.4,
        ytick={0.00, 0.05, 0.10, 0.15, 0.20, 0.25, 0.30, 0.35, 0.40},
        yticklabels={0.00, 0.05, 0.10, 0.15, 0.20, 0.25, 0.30, 0.35, 0.40},
        xtick={90, 91, 92, 93, 94, 95, 96, 97, 98, 99, 100},
        xticklabels={90, 91, 92, 93, 94, 95, 96, 97, 98, 99, 100},
        tick label style={font=\small},
        label style={font=\small},
        grid=major,
        grid style={dashed, gray!30},
        legend pos=north west, 
        legend cell align=left,
        legend style={font=\small, row sep=2pt},
        legend image post style={mark indices=},
        clip=false,
    ]

    \addplot[
        mark=*, thick, dashed, blue,
        mark size=2pt,
        mark indices={1,3,5,7,9,11,13,15, 17,19, 20}
    ] coordinates {
        (90.0, 0.03622) (90.5, 0.03866) (91.0, 0.0413) (91.5, 0.04427)
        (92.0, 0.04686) (92.5, 0.04985) (93.0, 0.05374) (93.5, 0.05787)
        (94.0, 0.06285) (94.5, 0.06833) (95.0, 0.07465) (95.5, 0.08231)
        (96.0, 0.09196) (96.5, 0.10082) (97.0, 0.11444) (97.5, 0.12867)
        (98.0, 0.1427)  (98.5, 0.17741) (99.0, 0.22253) (99.5, 0.33511)
    };
    \addlegendentry{AE}

    \addplot[
        mark=square*, thick, dashdotted, red,
        mark size=2pt,
        mark indices={1,3,5,7,9,11,13,15, 17,19, 20}
    ] coordinates {
        (90.0, 0.0868)  (90.5, 0.08832) (91.0, 0.09039) (91.5, 0.09246)
        (92.0, 0.09478) (92.5, 0.0974)  (93.0, 0.09977) (93.5, 0.1028)
        (94.0, 0.10596) (94.5, 0.10954) (95.0, 0.11196) (95.5, 0.11548)
        (96.0, 0.11913) (96.5, 0.12292) (97.0, 0.12723) (97.5, 0.13304)
        (98.0, 0.14223) (98.5, 0.15318) (99.0, 0.16807) (99.5, 0.19015)
    };
    \addlegendentry{HE}

    \addplot[
        mark=triangle*, thick, solid, myempirical,
        mark size=2.5pt, mark options={solid},
        mark indices={1,3,5,7,9,11,13,15, 17,19, 20}
    ] coordinates {
        (90.0, 0.04144) (90.5, 0.04349) (91.0, 0.0453)  (91.5, 0.04732)
        (92.0, 0.05006) (92.5, 0.05267) (93.0, 0.05546) (93.5, 0.05881)
        (94.0, 0.06312) (94.5, 0.06705) (95.0, 0.07073) (95.5, 0.07471)
        (96.0, 0.07984) (96.5, 0.08672) (97.0, 0.09498) (97.5, 0.10557)
        (98.0, 0.11505) (98.5, 0.13265) (99.0, 0.16262) (99.5, 0.19635)
    };
    \addlegendentry{Empirical}
    \end{axis}
\end{tikzpicture}
    \caption{\textbf{Regret (90th to 99.5th percentile).} AE has lower regret than HE across most percentiles but is more vulnerable to extreme regret. The empirical selector captures most of AE's gains while substantially reducing its tail risk.}
    \label{fig:regret_percentile}
\end{figure}

To visualize protection against costly mistakes, Figure~\ref{fig:regret_percentile} plots regret percentiles from the 90th to the 99.5th percentile. AE's advantage over HE persists through most of the range, but its regret rises sharply beyond the 98th percentile. HE protects us from such extreme regret. The empirical selector combines these strengths, capturing most of AE's gains while reducing its tail risk.

\begin{figure}[ht]
    \centering
    \begin{tikzpicture}
    \begin{axis}[
        width=12cm, 
        height=8cm,
        xlabel={SNR Decile},
        ylabel={Proportion of Signal ($S^2$)},
        ymin=0.60, ymax=0.92,
        ytick={0.60, 0.65, 0.70, 0.75, 0.80, 0.85, 0.90},
        yticklabels={0.60, 0.65, 0.70, 0.75, 0.80, 0.85, 0.90},
        xtick={0,...,9},
        xticklabels={1, 2, 3, 4, 5, 6, 7, 8, 9, 10},
        tick label style={font=\small},
        label style={font=\small},
        grid=major,
        grid style={dashed, gray!30},
        legend pos=south east,
        legend cell align=left,
        legend style={font=\small, row sep=2pt},
        clip=false,
    ]

    \addplot[
        mark=*, thick, dashed, blue,
        mark size=2pt
    ] coordinates {
        (0, 0.63905)
        (1, 0.73597)
        (2, 0.76585)
        (3, 0.80108)
        (4, 0.81402)
        (5, 0.83219)
        (6, 0.83186)
        (7, 0.83316)
        (8, 0.8587)
        (9, 0.89353)
    };
    \addlegendentry{AE}

    \addplot[
        mark=square*, thick, dashdotted, red,
        mark size=2pt
    ] coordinates {
        (0, 0.67358)
        (1, 0.73092)
        (2, 0.75284)
        (3, 0.77885)
        (4, 0.78469)
        (5, 0.79806)
        (6, 0.79408)
        (7, 0.79337)
        (8, 0.82311)
        (9, 0.86862)
    };
    \addlegendentry{HE}

    \addplot[
        mark=triangle*, thick, solid, myempirical,
        mark size=2.5pt, mark options={solid}
    ] coordinates {
        (0, 0.67336)
        (1, 0.74084)
        (2, 0.76849)
        (3, 0.79871)
        (4, 0.80976)
        (5, 0.82558)
        (6, 0.8273)
        (7, 0.82779)
        (8, 0.85537)
        (9, 0.89041)
    };
    \addlegendentry{Empirical}

    \addplot[
        mark=diamond*, thick, dotted, myoracle,
        mark size=2.5pt, mark options={solid}
    ] coordinates {
        (0, 0.70529)
        (1, 0.75901)
        (2, 0.78404)
        (3, 0.81078)
        (4, 0.82043)
        (5, 0.83606)
        (6, 0.83475)
        (7, 0.83557)
        (8, 0.86109)
        (9, 0.89481)
    };
    \addlegendentry{Oracle}

    \end{axis}
\end{tikzpicture}
    \caption{\textbf{Average $S^2$ by SNR decile for each selection strategy.} 
    AE outperforms HE in most deciles but falls behind in the lowest-SNR decile. 
    The empirical selector closely tracks the better-performing method across 
    deciles, nearly coinciding with the Oracle when SNR is high.}
    \label{fig:signal_by_decile}
\end{figure}

Figure~\ref{fig:signal_by_decile} breaks down average $S^2$ by SNR decile, where Decile~1 contains the weakest-signal datasets and Decile~10 the strongest. AE outperforms HE in most deciles, but falls behind in the lowest-SNR regime. This pattern reflects their complementary strengths: AE better exploits heterogeneity when the signal is strong, while HE is more reliable when it is weak. The empirical selector closely tracks the better-performing estimator across the full SNR range. It is close to AE in most deciles, nearly coincides with the oracle when SNR is high, and shifts toward HE in the weakest-signal regime.

We next quantify the practical cost of choosing honesty: how much additional data does HE require to match AE? For each dataset, we identify the smallest training size at which the better-performing method still significantly outperforms its alternative (at the 5\% level), re-tuning hyperparameters at each sample size so both methods are evaluated at their optimal complexity (see Appendix E for details).

Figure~\ref{fig:data_efficiency} shows the result averaged within each SNR decile. In the lowest-SNR decile, HE achieves comparable performance with roughly 7\% less data. As SNR increases, the cost of honesty grows, and HE requires up to 27\% more data to match AE. 

\begin{figure}[ht]
    \centering
    \begin{tikzpicture}
\begin{axis}[
    width=12cm,
    height=7cm,
    xlabel={SNR Decile},
    ylabel={Extra Data Honesty Needs},
    ylabel style={yshift=10pt},
    xtick={1,...,10},
    ymin=-10, ymax=30,
    ytick={-10, 0, 10, 20, 30},
    yticklabels={-10\%, 0\%, +10\%, +20\%, +30\%},
    grid=major,
    axis lines=left,
    enlargelimits=0.05,
    xticklabels={1, 2, 3, 4, 5, 6, 7, 8, 9, 10},
    tick label style={font=\small},
    label style={font=\small},
    title style={font=\small},
    legend style={at={(0.5,-0.22)}, anchor=north, font=\small},
    legend cell align=left,
    every axis plot/.append style={thick},
    clip=false,
]

\addplot[
    blue, solid, thick,
    mark=*, mark size=2pt
] coordinates {
    (1, -7.48)
    (2, -0.866)
    (3, 2.339)
    (4, 5.858)
    (5, 12.054)
    (6, 17.518)
    (7, 21.648)
    (8, 24.714)
    (9, 27.411)
    (10, 24.804)
};

\addplot[
    black, dashed, thick,
    forget plot
] coordinates {(1,0) (10,0)};

\end{axis}
\end{tikzpicture}
    \caption{\textbf{Additional data required for HE to match AE performance, by 
    SNR decile.} In the lowest-SNR regime, HE requires roughly 7\% less 
    data than AE, but up to 27\% more data as SNR increases.}
    \label{fig:data_efficiency}
\end{figure}

\subsection{Honesty as Regularization}

We now test the theoretical interpretation of honesty as regularization trading bias for variance. We estimate the bias--variance decomposition using repeated treatment randomization and train/test splits, following the randomization-based perspective of~\cite{athey2016econometrics}. Appendix F details the procedure.

One complication is that HE and AE do not operate at the same level of complexity. In our results, tuned AE forests tend to have fewer leaves than tuned HE forests, so comparing the two methods only at their separately tuned leaf sizes may confound estimator differences with differences in tuning.

To disentangle these effects, we match the two methods by number of leaves. Since HE uses only half the subsample for splitting, a minimum leaf size of $m$ for HE produces roughly the same number of leaves as an AE minimum leaf size of $2m$. We use this equivalence to define three regimes. In the \textbf{AE-matched} regime, AE uses its selected leaf size and HE uses half that value; in the \textbf{HE-matched} regime, HE uses its selected leaf size and AE uses twice that value. In the \textbf{self-optimal} regime, each method uses its own cross-validated leaf size. Figure~\ref{fig:bias_variance_tradeoff} summarizes results across SNR deciles, with all components standardized by $\mathrm{Var}(\beta)$.\footnote{Under this standardization, signal loss decomposes as $1-S^2=\text{Bias}^2+\text{Variance}$.}

\begin{figure}
    \centering
    \captionsetup[subfigure]{font=small}
    \begin{subfigure}[c]{0.49\textwidth}
        \centering
        \caption{AE-matched: Squared Bias}
        \begin{tikzpicture}
\begin{axis}[
    width=\linewidth,
    height=5.6cm,
    grid=major,
    grid style={dashed, gray!30},
    ymin=0, ymax=0.35,
    xtick={0,...,9},
    xticklabels={1, 2, 3, 4, 5, 6, 7, 8, 9, 10},
    ytick={0, 0.1, 0.2, 0.3},
    yticklabels={0, 0.1, 0.2, 0.3},
    tick label style={font=\scriptsize},
    label style={font=\small},
]
\addplot[mark=*, thick, blue, mark size=1.5pt] coordinates {
    (0, 0.16056)
    (1, 0.15181)
    (2, 0.14341)
    (3, 0.13693)
    (4, 0.14315)
    (5, 0.14093)
    (6, 0.15306)
    (7, 0.15707)
    (8, 0.1444)
    (9, 0.10439)
};
\addplot[mark=square*, thick, dashed, red, mark size=1.5pt] coordinates {
    (0, 0.30068)
    (1, 0.25916)
    (2, 0.23663)
    (3, 0.21784)
    (4, 0.21711)
    (5, 0.20818)
    (6, 0.21606)
    (7, 0.21709)
    (8, 0.1948)
    (9, 0.13844)
};
\end{axis}
\end{tikzpicture}
    \end{subfigure}
    \hfill
    \begin{subfigure}[c]{0.49\textwidth}
        \centering
        \caption{AE-matched: Variance}
        \begin{tikzpicture}
\begin{axis}[
    width=\linewidth,
    height=5.6cm,
    grid=major,
    grid style={dashed, gray!30},
    ymin=0, ymax=0.35,
    xtick={0,...,9},
    xticklabels={1, 2, 3, 4, 5, 6, 7, 8, 9, 10},
    ytick={0, 0.1, 0.2, 0.3},
    yticklabels={0, 0.1, 0.2, 0.3},
    tick label style={font=\scriptsize},
    label style={font=\small},
    legend style={at={(0.98,0.98)}, anchor=north east, font=\small},
]
\addplot[mark=*, thick, blue, mark size=1.5pt] coordinates {
    (0, 0.1958)
    (1, 0.1143)
    (2, 0.09042)
    (3, 0.06881)
    (4, 0.05492)
    (5, 0.04469)
    (6, 0.03616)
    (7, 0.03057)
    (8, 0.02391)
    (9, 0.0157)
};
\addlegendentry{AE}
\addplot[mark=square*, thick, dashed, red, mark size=1.5pt] coordinates {
    (0, 0.06208)
    (1, 0.03873)
    (2, 0.03189)
    (3, 0.02563)
    (4, 0.02159)
    (5, 0.01856)
    (6, 0.0159)
    (7, 0.01427)
    (8, 0.01184)
    (9, 0.00814)
};
\addlegendentry{HE}
\end{axis}
\end{tikzpicture}
    \end{subfigure}

    \begin{subfigure}[c]{0.49\textwidth}
        \centering
        \caption{Self-optimal: Squared Bias}
        \begin{tikzpicture}
\begin{axis}[
    width=\linewidth,
    height=5.6cm,
    grid=major,
    grid style={dashed, gray!30},
    ymin=0, ymax=0.35,
    xtick={0,...,9},
    xticklabels={1, 2, 3, 4, 5, 6, 7, 8, 9, 10},
    ytick={0, 0.1, 0.2, 0.3},
    yticklabels={0, 0.1, 0.2, 0.3},
    tick label style={font=\scriptsize},
    label style={font=\small},
]
\addplot[mark=*, thick, blue, mark size=1.5pt] coordinates {
    (0, 0.16056)
    (1, 0.15181)
    (2, 0.14341)
    (3, 0.13693)
    (4, 0.14315)
    (5, 0.14093)
    (6, 0.15306)
    (7, 0.15707)
    (8, 0.1444)
    (9, 0.10439)
};
\addplot[mark=square*, thick, dashed, red, mark size=1.5pt] coordinates {
    (0, 0.25324)
    (1, 0.22742)
    (2, 0.21519)
    (3, 0.20101)
    (4, 0.20628)
    (5, 0.20099)
    (6, 0.21161)
    (7, 0.21461)
    (8, 0.19206)
    (9, 0.13774)
};
\end{axis}
\end{tikzpicture}
    \end{subfigure}
    \hfill
    \begin{subfigure}[c]{0.49\textwidth}
        \centering
        \caption{Self-optimal: Variance}
        \begin{tikzpicture}
\begin{axis}[
    width=\linewidth,
    height=5.6cm,
    grid=major,
    grid style={dashed, gray!30},
    ymin=0, ymax=0.35,
    xtick={0,...,9},
    xticklabels={1, 2, 3, 4, 5, 6, 7, 8, 9, 10},
    ytick={0, 0.1, 0.2, 0.3},
    yticklabels={0, 0.1, 0.2, 0.3},
    tick label style={font=\scriptsize},
    label style={font=\small},
]
\addplot[mark=*, thick, blue, mark size=1.5pt] coordinates {
    (0, 0.1958)
    (1, 0.1143)
    (2, 0.09042)
    (3, 0.06881)
    (4, 0.05492)
    (5, 0.04469)
    (6, 0.03616)
    (7, 0.03057)
    (8, 0.02391)
    (9, 0.0157)
};
\addplot[mark=square*, thick, dashed, red, mark size=1.5pt] coordinates {
    (0, 0.08066)
    (1, 0.0474)
    (2, 0.03711)
    (3, 0.02882)
    (4, 0.02332)
    (5, 0.01953)
    (6, 0.01643)
    (7, 0.01453)
    (8, 0.01201)
    (9, 0.00823)
};
\end{axis}
\end{tikzpicture}
    \end{subfigure}

    \begin{subfigure}[c]{0.49\textwidth}
        \centering
        \caption{HE-matched: Squared Bias}
        \begin{tikzpicture}
\begin{axis}[
    width=\linewidth,
    height=5.6cm,
    grid=major,
    grid style={dashed, gray!30},
    ymin=0, ymax=0.35,
    xtick={0,...,9},
    xticklabels={1, 2, 3, 4, 5, 6, 7, 8, 9, 10},
    ytick={0, 0.1, 0.2, 0.3},
    yticklabels={0, 0.1, 0.2, 0.3},
    tick label style={font=\scriptsize},
    label style={font=\small},
    xlabel={SNR Decile},
]
\addplot[mark=*, thick, blue, mark size=1.5pt] coordinates {
    (0, 0.11912)
    (1, 0.12054)
    (2, 0.12118)
    (3, 0.11939)
    (4, 0.13171)
    (5, 0.13323)
    (6, 0.14828)
    (7, 0.15431)
    (8, 0.1412)
    (9, 0.10361)
};
\addplot[mark=square*, thick, dashed, red, mark size=1.5pt] coordinates {
    (0, 0.25324)
    (1, 0.22742)
    (2, 0.21519)
    (3, 0.20101)
    (4, 0.20628)
    (5, 0.20099)
    (6, 0.21161)
    (7, 0.21461)
    (8, 0.19206)
    (9, 0.13774)
};
\end{axis}
\end{tikzpicture}
    \end{subfigure}
    \hfill
    \begin{subfigure}[c]{0.49\textwidth}
        \centering
        \caption{HE-matched: Variance}
        \begin{tikzpicture}
\begin{axis}[
    width=\linewidth,
    height=5.6cm,
    grid=major,
    grid style={dashed, gray!30},
    ymin=0, ymax=0.35,
    xtick={0,...,9},
    xticklabels={1, 2, 3, 4, 5, 6, 7, 8, 9, 10},
    ytick={0, 0.1, 0.2, 0.3},
    yticklabels={0, 0.1, 0.2, 0.3},
    tick label style={font=\scriptsize},
    label style={font=\small},
    xlabel={SNR Decile},
]
\addplot[mark=*, thick, blue, mark size=1.5pt] coordinates {
    (0, 0.30274)
    (1, 0.16198)
    (2, 0.11907)
    (3, 0.08617)
    (4, 0.06432)
    (5, 0.04976)
    (6, 0.03894)
    (7, 0.03197)
    (8, 0.02472)
    (9, 0.01622)
};
\addplot[mark=square*, thick, dashed, red, mark size=1.5pt] coordinates {
    (0, 0.08066)
    (1, 0.0474)
    (2, 0.03711)
    (3, 0.02882)
    (4, 0.02332)
    (5, 0.01953)
    (6, 0.01643)
    (7, 0.01453)
    (8, 0.01201)
    (9, 0.00823)
};
\end{axis}
\end{tikzpicture}
    \end{subfigure}
    \caption{\textbf{Squared bias and variance across SNR deciles.} AE has lower bias, while HE has lower variance, across all regimes. AE outperforms HE in high-SNR settings because the variance gap closes faster than the bias gap as SNR increases.}
    \label{fig:bias_variance_tradeoff}
\end{figure}

The pattern is clean. AE has lower bias and HE has lower variance across all three regimes, with both gaps widest at low SNR and narrowing as signal strengthens. The variance gap closes faster, making AE the more likely winner as SNR increases. The results confirm the theory: honesty buys stability at the cost of fit, which can hurt accuracy when signal is strong.

\begin{figure}[ht]
    \centering
    \captionsetup[subfigure]{font=small}
    \begin{subfigure}[c]{0.365\textwidth}
    \caption{Sq.\ Approximation Bias}
    \begin{tikzpicture}
\begin{axis}[
    width=\linewidth,
    height=5.6cm,
    grid=major,
    grid style={dashed, gray!30},
    ymin=-0.12, ymax=0.3,
    xtick={0,1,2,3,4,5,6,7,8,9},
    xticklabels={1,2,3,4,5,6,7,8,9,10},
    ytick={-0.1, 0, 0.1, 0.2, 0.3},
    yticklabels={-0.1, 0.0, 0.1, 0.2, 0.3},
    tick label style={font=\scriptsize},
    label style={font=\footnotesize},
]
\addplot[mark=*, thick, blue, mark size=1.0pt] coordinates {
    (0,0.22778)(1,0.20682)(2,0.19185)(3,0.17828)(4,0.18018)
    (5,0.17462)(6,0.18401)(7,0.18601)(8,0.16849)(9,0.11996)
};
\addplot[mark=square*, thick, dashed, red, mark size=1.0pt] coordinates {
    (0,0.24900)(1,0.22413)(2,0.21222)(3,0.19816)(4,0.20352)
    (5,0.19862)(6,0.20915)(7,0.21237)(8,0.19002)(9,0.13632)
};
\addplot[black, thin, dashed, forget plot] coordinates {(0,0) (9,0)};
\end{axis}
\end{tikzpicture}
    \end{subfigure}
    \hspace{-0.06\textwidth}
    \begin{subfigure}[c]{0.365\textwidth}
    \caption{Sq.\ Estimation Bias}
    \begin{tikzpicture}
\begin{axis}[
    width=\linewidth,
    height=5.6cm,
    grid=major,
    grid style={dashed, gray!30},
    ymin=-0.12, ymax=0.3,
    xtick={0,1,2,3,4,5,6,7,8,9},
    xticklabels={1,2,3,4,5,6,7,8,9,10},
    ytick={-0.1, 0, 0.1, 0.2, 0.3},
    yticklabels={-0.1, 0.0, 0.1, 0.2, 0.3},
    tick label style={font=\scriptsize},
    label style={font=\footnotesize},
]
\addplot[mark=*, thick, blue, mark size=1.0pt] coordinates {
    (0,0.03442)(1,0.02467)(2,0.02074)(3,0.01636)(4,0.01379)
    (5,0.01205)(6,0.01011)(7,0.00887)(8,0.00698)(9,0.00445)
};
\addplot[mark=square*, thick, dashed, red, mark size=1.0pt] coordinates {
    (0,0.00141)(1,0.00079)(2,0.00061)(3,0.00048)(4,0.00041)
    (5,0.00035)(6,0.00031)(7,0.00028)(8,0.00025)(9,0.00022)
};
\addplot[black, thin, dashed, forget plot] coordinates {(0,0) (9,0)};
\end{axis}
\end{tikzpicture}
    \end{subfigure}
    \hspace{-0.06\textwidth}
    \begin{subfigure}[c]{0.365\textwidth}
    \caption{Bias Interaction}
    \begin{tikzpicture}
\begin{axis}[
    width=\linewidth,
    height=5.6cm,
    grid=major,
    grid style={dashed, gray!30},
    ymin=-0.12, ymax=0.3,
    xtick={0,1,2,3,4,5,6,7,8,9},
    xticklabels={1,2,3,4,5,6,7,8,9,10},
    ytick={-0.1, 0, 0.1, 0.2, 0.3},
    yticklabels={-0.1, 0.0, 0.1, 0.2, 0.3},
    tick label style={font=\scriptsize},
    label style={font=\footnotesize},
    legend style={font=\scriptsize, inner sep=1.0pt}
]
\addplot[mark=*, thick, blue, mark size=1.0pt] coordinates {
    (0,-0.10165)(1,-0.07969)(2,-0.06918)(3,-0.05770)(4,-0.05082)
    (5,-0.04575)(6,-0.04105)(7,-0.03781)(8,-0.03107)(9,-0.02001)
};
\addlegendentry{AE}
\addplot[mark=square*, thick, dashed, red, mark size=1.0pt] coordinates {
    (0,0.00282)(1,0.00250)(2,0.00235)(3,0.00236)(4,0.00235)
    (5,0.00203)(6,0.00215)(7,0.00196)(8,0.00178)(9,0.00120)
};
\addlegendentry{HE}
\addplot[black, thin, dashed, forget plot] coordinates {(0,0) (9,0)};
\end{axis}
\end{tikzpicture}
    \end{subfigure}

    \begin{subfigure}[c]{0.365\textwidth}
    \caption{Target Coupling}
    \begin{tikzpicture}
\begin{axis}[
    width=\linewidth,
    height=5.6cm,
    grid=major,
    grid style={dashed, gray!30},
    ymin=-0.12, ymax=0.3,
    xtick={0,1,2,3,4,5,6,7,8,9},
    xticklabels={1,2,3,4,5,6,7,8,9,10},
    ytick={-0.1, 0, 0.1, 0.2, 0.3},
    yticklabels={-0.1, 0.0, 0.1, 0.2, 0.3},
    tick label style={font=\scriptsize},
    label style={font=\footnotesize},
    xlabel={SNR Decile},
]
\addplot[mark=*, thick, blue, mark size=1.0pt] coordinates {
    (0,0.02110)(1,0.01839)(2,0.01689)(3,0.01611)(4,0.01547)
    (5,0.01527)(6,0.01544)(7,0.01562)(8,0.01440)(9,0.01234)
};
\addplot[mark=square*, thick, dashed, red, mark size=1.0pt] coordinates {
    (0,0.01150)(1,0.01033)(2,0.00983)(3,0.00948)(4,0.00925)
    (5,0.00925)(6,0.00924)(7,0.00943)(8,0.00886)(9,0.00799)
};
\addplot[black, thin, dashed, forget plot] coordinates {(0,0) (9,0)};
\end{axis}
\end{tikzpicture}
    \end{subfigure}
    \hspace{-0.06\textwidth}
    \begin{subfigure}[c]{0.365\textwidth}
    \caption{Noise Overlap}
    \begin{tikzpicture}
\begin{axis}[
    width=\linewidth,
    height=5.6cm,
    grid=major,
    grid style={dashed, gray!30},
    ymin=-0.12, ymax=0.3,
    xtick={0,1,2,3,4,5,6,7,8,9},
    xticklabels={1,2,3,4,5,6,7,8,9,10},
    ytick={-0.1, 0, 0.1, 0.2, 0.3},
    yticklabels={-0.1, 0.0, 0.1, 0.2, 0.3},
    tick label style={font=\scriptsize},
    label style={font=\footnotesize},
    xlabel={SNR Decile},
]
\addplot[mark=*, thick, blue, mark size=1.0pt] coordinates {
    (0,0.15484)(1,0.08442)(2,0.06649)(3,0.04973)(4,0.03956)
    (5,0.03242)(6,0.02709)(7,0.02361)(8,0.01939)(9,0.01592)
};
\addplot[mark=square*, thick, dashed, red, mark size=1.0pt] coordinates {
    (0,0.06244)(1,0.03434)(2,0.02645)(3,0.02030)(4,0.01643)
    (5,0.01403)(6,0.01244)(7,0.01159)(8,0.01029)(9,0.00944)
};
\addplot[black, thin, dashed, forget plot] coordinates {(0,0) (9,0)};
\end{axis}
\end{tikzpicture}
    \end{subfigure}
    \hspace{-0.06\textwidth}
    \begin{subfigure}[c]{0.365\textwidth}
    \caption{Target--Est. Spillover}
    \begin{tikzpicture}
\begin{axis}[
    width=\linewidth,
    height=5.6cm,
    grid=major,
    grid style={dashed, gray!30},
    ymin=-0.12, ymax=0.3,
    xtick={0,1,2,3,4,5,6,7,8,9},
    xticklabels={1,2,3,4,5,6,7,8,9,10},
    ytick={-0.1, 0, 0.1, 0.2, 0.3},
    yticklabels={-0.1, 0.0, 0.1, 0.2, 0.3},
    tick label style={font=\scriptsize},
    label style={font=\footnotesize},
    xlabel={SNR Decile},
]
\addplot[mark=*, thick, blue, mark size=1.0pt] coordinates {
    (0,0.01986)(1,0.01148)(2,0.00704)(3,0.00297)(4,-0.00011)
    (5,-0.00301)(6,-0.00637)(7,-0.00866)(8,-0.00987)(9,-0.01256)
};
\addplot[mark=square*, thick, dashed, red, mark size=1.0pt] coordinates {
    (0,0.00673)(1,0.00273)(2,0.00084)(3,-0.00096)(4,-0.00237)
    (5,-0.00374)(6,-0.00525)(7,-0.00648)(8,-0.00714)(9,-0.00920)
};
\addplot[black, thin, dashed, forget plot] coordinates {(0,0) (9,0)};
\end{axis}
\end{tikzpicture}
    \end{subfigure}
    \caption{\textbf{Bias--variance subcomponents across SNR deciles.} 
    Squared bias is dominated by approximation bias, while variance is dominated by noise overlap.}
    \label{fig:bias_variance_subcomponents}
\end{figure}

Figure~\ref{fig:bias_variance_subcomponents} further decomposes these quantities for the self-optimal regime. Squared bias is the sum of squared approximation bias, squared estimation bias, and their interaction. Variance is the sum of target coupling, noise overlap, and target--estimation spillover.  

On the bias side, approximation bias dominates, and AE reduces it as theory predicts. Estimation bias is small for AE and zero for HE. Interestingly, the bias interaction for AE is \emph{negative}: estimation bias actually reduces overall bias because the two components tend to have opposite signs. This is consistent with the stylized example in Section~\ref{sec:example}, where AE has negative approximation bias and positive estimation bias.  

On the variance side, noise overlap accounts for most of the gap, and HE reduces it as theory predicts. Theory also predicts higher target coupling for HE at high SNR, but we do not observe it, suggesting the SNR threshold for that is too high to be practically relevant. Target--estimation spillover is generally closer to zero for HE, consistent with HE weakening the link between split selection and effect estimation.\footnote{Because target effects are estimated from a finite auxiliary sample, the estimated spillover is shifted downward and can be negative even when the population quantity is positive. Its sign alone is therefore not informative; see Appendix~F.}

\section{Discussion and Broader Implications}\label{sec:discussion}

Honesty is widely treated as a conservative default in causal forests: it is built into common software, frequently recommended, and often adopted without much deliberation. Our results challenge this convention. Honesty is not inherently desirable and often degrades performance when treatment effect heterogeneity is substantial and the data large enough. In our empirical study, defaulting to honesty can require up to 27\% more training data to match the performance of forests trained without it.

These findings have immediate implications for applied workflows that use causal forests for individual-level targeting, personalization, and prioritization.
\begin{enumerate}
    \item \textbf{Treat honesty as a hyperparameter.} For the goal studied here---accurate CATE prediction at the individual level---there is little justification for adopting HE reflexively. Practitioners should treat honesty as a modeling choice, just like other complexity choices such as tree depth.
    \item \textbf{Adaptive estimation is a more sensible default.} Most software packages default to honest estimation, but our results suggest the opposite default is more defensible. AE outperforms HE in most settings. In the cases where HE might seem preferable---when SNR is low and overfitting is a concern---the signal may be too weak for causal forests to reliably model heterogeneity at all. We return to this point below.
    \item \textbf{Choosing honesty often implies greater data requirements.} When experiments are costly or samples are limited, the possibility that honesty may demand substantially more data to reach a given level of CATE accuracy should be part of the design conversation.
\end{enumerate}

In our empirical analysis, we defined SNR as a property of the data-generating process, but it is better understood as a sample-dependent quantity. More data makes effect heterogeneity easier to model, increasing the effective SNR and reducing the regularization benefit of honesty. AE tends to gain an advantage as sample size grows, but with sufficiently large samples both approaches perform well and practical differences diminish. Appendix H traces this progression empirically, showing the SNR-based insight holds even at hundreds of thousands of observations.

Honesty may still be preferable when the goal extends beyond predictive accuracy. Theoretical guarantees support HE for tasks like constructing confidence intervals or testing hypotheses, even at some cost to predictive accuracy. However, point estimation and uncertainty quantification do not need to rely on the same model. A practitioner could use an honest forest for testing or interval construction, and a separate adaptive forest for individual-level estimates used in targeting or personalization. The key is to match the estimation strategy to the objective rather than treating honesty as a universal default.

That said, causal forests may not even be the right tool in the first place. Before choosing between AE and HE, practitioners should assess whether causal forests are appropriate for their setting. Interestingly, the conditions that favor AE over HE may also be those where causal forests are most justified: high SNR implies substantial, detectable heterogeneity---precisely the regime where a flexible nonparametric learner gains an advantage over simpler alternatives. Conversely, when SNR is low and HE might seem preferable, the signal may be too weak for causal forests to reliably model heterogeneity at all. Indeed, in our empirical analysis, we find that the datasets where neither AE nor HE pass the heterogeneity test cluster heavily in low-SNR regimes. This suggests that when the signal is weak, the choice between AE and HE may matter less than the choice of whether to use causal forests at all.

The tension we study applies whenever the model structure is chosen using the same data later used for estimation---whether in tree-based methods, variable selection procedures, or ensemble learners. Separating selection from estimation reduces overfitting but limits flexibility to capture heterogeneity, and whether that trade-off is desirable depends on signal strength, sample size, and the analysis objective. Appendix G illustrates the same tension using adaptive and honest Lasso implementations on the ACIC datasets. At the same time, the specific decomposition components we analyze are closely tied to recursive partitioning methods and would likely manifest differently in other estimators. Systematically characterizing analogous components across CATE estimators is an interesting avenue for future research.

This paper highlights a distinction that is central to causal machine learning: the separation between approximation error and estimation error. In our setting, approximation error arises because the object we estimate---the SPATE associated with a leaf---is only an approximation to the individual-level CATE. Estimation error arises from sampling variability and selection bias in estimating those leaf-level effects. Honesty shifts the balance between these two sources of error.

A similar structural tension appears throughout causal inference. The statistical target---what can be identified from data, such as a conditional difference in means---need not equal the causal target, such as the CATE. In the presence of confounding, this mismatch is a form of approximation error: the estimable quantity does not equal the quantity of interest.

Identification assumptions formalize when this gap disappears. When they hold, approximation error vanishes, but finite-sample considerations remain. Choices that reduce approximation error, such as stronger identification strategies, can increase estimation error. A method that aggressively corrects for confounding bias may substantially increase variance, resulting in larger prediction error. The same tension appears when comparing randomized experiments to large observational datasets: experiments eliminate approximation error due to confounding, while observational data reduces estimation error through scale~\citep{fernandez2025observational}. Aligning the statistical and causal targets is ultimately a choice about how to navigate the bias--variance trade-off, much like honesty is in our setting.

The broader lesson is that causal analysis often involves trade-offs between approximation and estimation error. Our contribution is to make this trade-off explicit in heterogeneous treatment effect estimation with forests. Honesty is not a universal safeguard but a design choice that shapes this trade-off. More generally, we hope this perspective encourages researchers to ask, for any design choice in causal machine learning: how does it shift the balance between approximation and estimation error, and under what conditions does that shift help or hurt?



\section*{Acknowledgments}

We dedicate this work to the memory of Sriram Somanchi. Conversations with Sriram during a visit to Dagstuhl convinced us that these findings were worth pursuing as a paper. They also helped shape the early direction of our research. His intellectual generosity, warmth, and curiosity were an inspiration to us, and he is deeply missed.

We thank Foster Provost for many insightful discussions and constructive feedback throughout the development of this work. We are grateful to our research assistant, Pauline (Pui Lam) Luk, for doing much of the heavy lifting in organizing and synthesizing the literature review.

\bibliographystyle{plainnat}
\bibliography{reference}

@article{wager2018estimation,
  title={Estimation and inference of heterogeneous treatment effects using random forests},
  author={Wager, Stefan and Athey, Susan},
  journal={Journal of the American Statistical Association},
  volume={113},
  number={523},
  pages={1228--1242},
  year={2018},
  publisher={Taylor \& Francis}
}

@article{athey2016recursive,
  title={Recursive partitioning for heterogeneous causal effects},
  author={Athey, Susan and Imbens, Guido},
  journal={Proceedings of the National Academy of Sciences},
  volume={113},
  number={27},
  pages={7353--7360},
  year={2016},
  publisher={National Academy of Sciences}
}

@article{athey2019generalized,
  title = {Generalized Random Forests},
  author = {Athey, Susan and Tibshirani, Julie and Wager, Stefan},
  journal = {The Annals of Statistics},
  volume = {47},
  number = {2},
  pages = {1148--1178},
  year = {2019}
}

@article{dorie2019automated,
  title={Automated versus Do-It-Yourself Methods for Causal Inference: Lessons Learned from a Data Analysis Competition},
  author={Dorie, Vincent and Hill, Jennifer and Shalit, Uri and Scott, Marc and Cervone, Daniel},
  journal={Statistical science},
  volume={34},
  number={1},
  pages={43--68},
  year={2019},
  publisher={Institute of Mathematical Statistics}
}

@misc{econml,
  author={Keith Battocchi and Eleanor Dillon and Maggie Hei and Greg Lewis and Paul Oka and Miruna Oprescu and Vasilis Syrgkanis},
  title={{EconML}: {A Python Package for ML-Based Heterogeneous Treatment Effects Estimation}},
  howpublished={https://github.com/py-why/EconML},
  note={Version 0.x},
  year={2019}
}

@article{athey2015machine,
  title={Machine Learning Methods for Estimating Heterogeneous Causal Effects},
  author={Athey, Susan and Imbens, Guido W.},
  journal={arXiv preprint arXiv:1504.01132v1},
  year={2015}
}

@article{imai2025statistical,
  title={Statistical inference for heterogeneous treatment effects discovered by generic machine learning in randomized experiments},
  author={Imai, Kosuke and Li, Michael Lingzhi},
  journal={Journal of Business \& Economic Statistics},
  volume={43},
  number={1},
  pages={256--268},
  year={2025},
  publisher={Taylor \& Francis}
}

@article{athey2016econometrics,
  title={The econometrics of randomized experiments},
  author={Athey, Susan and Imbens, Guido W.},
  journal={arXiv preprint arXiv:1607.00698},
  year={2016}
}

@article{scornet2016asymptotics,
  title={On the asymptotics of random forests},
  author={Scornet, Erwan},
  journal={Journal of Multivariate Analysis},
  volume={146},
  pages={72--83},
  year={2016},
  publisher={Elsevier}
}

@inproceedings{klusowski2021sharp,
  author    = {Klusowski, Jason},
  title     = {Sharp analysis of a simple model for random forests},
  editor    = {Banerjee, Arindam and Fukumizu, Kenji},
  booktitle = {Proceedings of The 24th International Conference on Artificial Intelligence and Statistics},
  series    = {Proceedings of Machine Learning Research},
  volume    = {130},
  pages     = {757--765},
  publisher = {PMLR},
  year      = {2021}
}

@book{imbens2015causal,
  title={Causal Inference for Statistics, Social, and Biomedical Sciences: An Introduction},
  author={Imbens, Guido W. and Rubin, Donald B.},
  year={2015},
  publisher={Cambridge University Press},
  address={Cambridge, UK}
}

@article{fernandez2025observational,
  title={Observational vs. Experimental Data When Making Automated Decisions Using Machine Learning},
  author={Fern{\'a}ndez-Lor{\'\i}a, Carlos and Provost, Foster},
  journal={INFORMS Journal on Data Science},
  volume = {4},
  number = {3},
  pages = {197-229},
  year = {2025},
  publisher={INFORMS}
}

@article{bladt2026consistency,
  title={Consistency of Honest Decision Trees and Random Forests},
  author={Bladt, Martin and Lemvig, Rasmus Frigaard},
  journal={arXiv preprint arXiv:2601.14991},
  year={2026}
}

@article{cattaneo2025honest,
  title={The honest truth about causal trees: Accuracy limits for heterogeneous treatment effect estimation},
  author={Cattaneo, Matias D and Klusowski, Jason M and Yu, Ruiqi Rae},
  journal={arXiv preprint arXiv:2509.11381},
  year={2025}
}

@article{cattaneo2022pointwise,
  title={On the pointwise behavior of recursive partitioning and its implications for heterogeneous causal effect estimation},
  author={Cattaneo, Matias D and Klusowski, Jason M and Tian, Peter M},
  journal={arXiv preprint arXiv:2211.10805},
  year={2022}
}

@misc{havelka2022honesty,
  author={Havelka, Matej},
  title={Honesty in Causal Forests, is it worth it?},
  note={Bachelor's thesis, Delft University of Technology},
  year={2022},
  howpublished={https://repository.tudelft.nl/file/File\_86d8e7a1-8951-4b8a-8714-63a01ba9c805}
}

@misc{prodan2025analyzing,
  author={Prodan, Rheea-Maria},
  title={Analyzing the Impact of Depth and Leaf Size on CATE Estimation in Honest Causal Trees},
  note={Bachelor's thesis, Delft University of Technology},
  year={2025},
  howpublished={https://repository.tudelft.nl/file/File\_edf8b751-f408-4186-b343-6bcfb2aa6765}
}

@article{kallus2023stochastic,
  title={Stochastic optimization forests},
  author={Kallus, Nathan and Mao, Xiaojie},
  journal={Management Science},
  volume={69},
  number={4},
  pages={1975--1994},
  year={2023},
  publisher={INFORMS}
}

@book{breiman1984classification,
  title     = {Classification and Regression Trees},
  author    = {Breiman, Leo and Friedman, Jerome H. and Olshen, Richard A. and Stone, Charles J.},
  year      = {1984},
  publisher = {Wadsworth International Group},
  address   = {Belmont, CA}
}

@book{hastie2009elements,
  author    = {Hastie, Trevor and Tibshirani, Robert and Friedman, Jerome},
  title     = {The Elements of Statistical Learning: Data Mining, Inference, and Prediction},
  edition   = {2nd},
  year      = {2009},
  publisher = {Springer},
  address   = {New York},
}

@article{dietterich1998approximate,
  title={Approximate statistical tests for comparing supervised classification learning algorithms},
  author={Dietterich, Thomas G.},
  journal={Neural computation},
  volume={10},
  number={7},
  pages={1895--1923},
  year={1998},
  publisher={MIT Press}
}

@article{friedman2010regularization,
  title={Regularization paths for generalized linear models via coordinate descent},
  author={Friedman, Jerome H. and Hastie, Trevor and Tibshirani, Rob},
  journal={Journal of statistical software},
  volume={33},
  number={1},
  pages={1--22},
  year={2010}
}

@article{lei2020cross,
  title={Cross-validation with confidence},
  author={Lei, Jing},
  journal={Journal of the American Statistical Association},
  volume={115},
  number={532},
  pages={1978--1997},
  year={2020},
  publisher={Taylor \& Francis}
}

@article{hammond2024identifying,
  title={Identifying and fitting eclipse maps of exoplanets with cross-validation},
  author={Hammond, Mark and Lewis, Neil T and Boone, Sasha and Chen, Xueqing and Mendon{\c{c}}a, Jo{\~a}o M and Parmentier, Vivien and Taylor, Jake and Bell, Taylor and dos Santos, Leonardo and Crouzet, Nicolas and others},
  journal={Monthly Notices of the Royal Astronomical Society},
  volume={532},
  number={4},
  pages={4350--4368},
  year={2024},
  publisher={Oxford University Press}
}

@article{piironen2020projective,
  title={Projective inference in high-dimensional problems: Prediction and feature selection},
  author={Piironen, Juho and Paasiniemi, Markus and Vehtari, Aki},
  journal={Electronic Journal of Statistics},
  volume={14},
  number={1},
  pages={2155--2197},
  year={2020},
  publisher={Institute of Mathematical Statistics}
}

@article{villanueva2024system,
  title={System-wide analysis of RNA and protein subcellular localization dynamics},
  author={Villanueva, Eneko and Smith, Tom and Pizzinga, Mariavittoria and Elzek, Mohamed and Queiroz, Rayner ML and Harvey, Robert F and Breckels, Lisa M and Crook, Oliver M and Monti, Mie and Dezi, Veronica and others},
  journal={Nature Methods},
  volume={21},
  number={1},
  pages={60--71},
  year={2024},
  publisher={Nature Publishing Group US New York}
}

@article{windmeijer2019use,
  title={On the use of the lasso for instrumental variables estimation with some invalid instruments},
  author={Windmeijer, Frank and Farbmacher, Helmut and Davies, Neil and Davey Smith, George},
  journal={Journal of the American Statistical Association},
  volume={114},
  number={527},
  pages={1339--1350},
  year={2019},
  publisher={Taylor \& Francis}
}

\clearpage
\appendix
\section{Splitting Criterion}\label{app:splitting_criterion}

\noindent
\textbf{Proposition.} For any fixed candidate split, the splitting criterion in Equation~\eqref{eq:split_criterion} is a consistent estimator of the MSE reduction presented in Equation~\eqref{eq:mse_gain}.

\smallskip
\noindent
\textit{Proof.} The node MSE in Equation~\eqref{eq:node_mse} is:
\begin{align}
\text{MSE}(\ell)
&= \mathbb{E} \left[ \left( \beta(\mathbf{X}) - \tau(\ell) \right)^2 \mid \mathbf{X} \in \ell \right] \\
&= \mathbb{E}[\beta(\mathbf{X})^2 \mid \mathbf{X} \in \ell]
- 2\tau(\ell) \cdot \mathbb{E}[\beta(\mathbf{X}) \mid \mathbf{X} \in \ell]
+ \tau(\ell)^2 \\
&= \mathbb{E}[\beta(\mathbf{X})^2 \mid \mathbf{X} \in \ell] - \tau(\ell)^2,
\end{align}
where we use the fact that $\tau(\ell) = \mathbb{E}[\beta(\mathbf{X}) \mid \mathbf{X} \in \ell]$.

\smallskip
\noindent
Now consider a split of a parent node $\ell_0$ into child nodes $\ell_1$ and $\ell_2$. Throughout, $\mathbb{P}(\mathbf{X} \in \ell_i)$ abbreviates $\mathbb{P}(\mathbf{X} \in \ell_i \mid \mathbf{X} \in \ell_0)$, so that $\mathbb{P}(\mathbf{X} \in \ell_1) + \mathbb{P}(\mathbf{X} \in \ell_2) = 1$. The reduction in MSE is:
\begin{align}
\text{Reduction in MSE}
&= \text{MSE}(\ell_0) 
- \mathbb{P}(\mathbf{X} \in \ell_1) \cdot \text{MSE}(\ell_1)
- \mathbb{P}(\mathbf{X} \in \ell_2) \cdot \text{MSE}(\ell_2) \\
&= \mathbb{E}[\beta(\mathbf{X})^2 \mid \mathbf{X} \in \ell_0] - \tau(\ell_0)^2 \\
&\quad - \mathbb{P}(\mathbf{X} \in \ell_1) \left( \mathbb{E}[\beta(\mathbf{X})^2 \mid \mathbf{X} \in \ell_1] - \tau(\ell_1)^2 \right) \nonumber \\
&\quad - \mathbb{P}(\mathbf{X} \in \ell_2) \left( \mathbb{E}[\beta(\mathbf{X})^2 \mid \mathbf{X} \in \ell_2] - \tau(\ell_2)^2 \right) \nonumber \\
&= -\tau(\ell_0)^2 
+ \mathbb{P}(\mathbf{X} \in \ell_1) \cdot \tau(\ell_1)^2 
+ \mathbb{P}(\mathbf{X} \in \ell_2) \cdot \tau(\ell_2)^2.
\end{align}
\smallskip
\noindent
Substitute $\tau(\ell_0) = \mathbb{P}(\mathbf{X} \in \ell_1) \cdot \tau(\ell_1) + \mathbb{P}(\mathbf{X} \in \ell_2) \cdot \tau(\ell_2)$ into the expression:
\begin{align}
\text{Reduction in MSE}
&= \mathbb{P}(\mathbf{X} \in \ell_1) \cdot \tau(\ell_1)^2 
+ \mathbb{P}(\mathbf{X} \in \ell_2) \cdot \tau(\ell_2)^2 \\
&\quad - \left( \mathbb{P}(\mathbf{X} \in \ell_1) \cdot \tau(\ell_1)
+ \mathbb{P}(\mathbf{X} \in \ell_2) \cdot \tau(\ell_2) \right)^2 \nonumber \\
&= \mathbb{P}(\mathbf{X} \in \ell_1) (1 - \mathbb{P}(\mathbf{X} \in \ell_1)) \cdot \tau(\ell_1)^2 \\
&\quad + \mathbb{P}(\mathbf{X} \in \ell_2) (1 - \mathbb{P}(\mathbf{X} \in \ell_2)) \cdot \tau(\ell_2)^2 \nonumber \\
&\quad - 2 \mathbb{P}(\mathbf{X} \in \ell_1) \mathbb{P}(\mathbf{X} \in \ell_2) \cdot \tau(\ell_1) \tau(\ell_2) \nonumber \\
&= \mathbb{P}(\mathbf{X} \in \ell_1) \mathbb{P}(\mathbf{X} \in \ell_2) 
\cdot \left( \tau(\ell_1) - \tau(\ell_2) \right)^2.
\end{align}
\smallskip
\noindent
Under ignorability, $\hat\tau(\ell_i;\mathcal{S}_{\text{sp}})$ is consistent for $\tau(\ell_i)$ and $n_{\ell_i}/n_{\ell_0}$ for $\mathbb{P}(\mathbf{X}\in\ell_i\mid\mathbf{X}\in\ell_0)$, so by the law of large numbers and the continuous mapping theorem the sample analog
\begin{equation}
\text{Split criterion} = 
\frac{n_{\ell_1} n_{\ell_2}}{(n_{\ell_1} +n_{\ell_2})^2}
\left( \hat\tau(\ell_1; \mathcal{S}_{\text{sp}}) - \hat\tau(\ell_2; \mathcal{S}_{\text{sp}}) \right)^2
\end{equation}
converges in probability to the reduction in MSE as $n_{\ell_0}$ grows.

\section{Proofs for Analytical Example} \label{app:example_proofs}

All subsequent proofs pertain to adaptive estimation (AE). Therefore, for simplicity in notation, we will use $\hat\beta$ and $\hat\ell$ instead of $\hat\beta^{\text{A}}$ and $\hat\ell^{\text{A}}$.

\subsection{Common Setup and Notation}\label{app:setup}

Let $X_j \sim \text{Bern}(0.5)$ for $j=1,\cdots,J$. We assume that $X_1$ is informative about treatment effects, while the remaining features are uninformative about the potential outcomes, and that
\begin{equation}
    X_1\perp\{X_j\}_{j\neq 1}.
\end{equation}
Treatment is randomly assigned with $\mathbb P(T=1)=0.5$. We also assume homoscedasticity:
\begin{equation}
    \mathrm{Var}(Y\mid\mathbf{X},T)=\sigma^2.
\end{equation}

Splitting on $X_j$ yields estimated SPATEs
\begin{equation}
\hat\tau_j(0), \hat\tau_j(1),
\end{equation}
defined as the difference in mean outcomes between treated and control units among individuals with \( X_j = 0 \) and \( X_j = 1 \), respectively, using the full training sample $\mathcal{S}_{\text{tr}}$.

Define:
\begin{align}
\Delta_j &:= \hat\tau_j(1) - \hat\tau_j(0), \label{eq:delta_def} \\
\eta_j &:= \hat\tau_j(1) + \hat\tau_j(0), \label{eq:eta_def}
\end{align}
so that:
\begin{align}
\hat\tau_j(1) &= \frac{1}{2}(\eta_j + \Delta_j), \label{eq:tau1_decomp} \\
\hat\tau_j(0) &= \frac{1}{2}(\eta_j - \Delta_j). \label{eq:tau0_decomp}
\end{align}

Under random treatment assignment, the difference-in-means estimators are unbiased for the corresponding SPATEs. For sufficiently large samples, applying the multivariate Central Limit Theorem jointly to
\[
\bigl(\hat\tau_1(0),\hat\tau_1(1),\ldots,\hat\tau_J(0),\hat\tau_J(1)\bigr)
\]
implies that this vector is approximately jointly normal. Because $\eta_j$ and $\Delta_j$ are linear combinations of the SPATE estimators, the vector containing all $\eta_j$'s and $\Delta_j$'s is also approximately jointly normal.

Under this joint-normal approximation, the feature-specific marginal distributions depend on whether $X_j$ is informative. Here, $\sigma_j^2$ denotes the sampling variance of either child-node SPATE estimator for feature $j$.

\begin{itemize}
    \item \textbf{Informative feature} \( (j = 1) \): \(\tau_1(1) = \theta\), \(\tau_1(0) = -\theta\), so
    \begin{align}
        \hat\tau_1(1) &\sim \mathcal{N}(\theta, \sigma^2_1), \quad 
        \hat\tau_1(0) \sim \mathcal{N}(-\theta, \sigma^2_1), \\
        \Delta_1 &\sim \mathcal{N}(2\theta, 2\sigma^2_1), \quad 
        \eta_1 \sim \mathcal{N}(0, 2\sigma^2_1).
    \end{align}
    \item \textbf{Uninformative feature} \( (j > 1) \): \(\tau_j(0) = \tau_j(1) = 0\), so
    \begin{align}
        \hat\tau_j(0), \hat\tau_j(1) &\sim \mathcal{N}(0, \sigma^2_j), \\
        \Delta_j, \eta_j &\sim \mathcal{N}(0, 2\sigma^2_j), \\
        \sigma^2_j > \sigma^2_1 & \text{  as a result of ignoring heterogeneity explained by $X_1$.}
    \end{align}
\end{itemize}

We next characterize the dependence among these quantities. For any features $j$ and $l$,
\begin{equation}
\begin{aligned}
\operatorname{Cov}(\eta_j,\Delta_l)
&=
\operatorname{Cov}\bigl(\hat\tau_j(1),\hat\tau_l(1)\bigr)
+\operatorname{Cov}\bigl(\hat\tau_j(0),\hat\tau_l(1)\bigr)\\
&\quad-
\operatorname{Cov}\bigl(\hat\tau_j(1),\hat\tau_l(0)\bigr)
-\operatorname{Cov}\bigl(\hat\tau_j(0),\hat\tau_l(0)\bigr).
\end{aligned}
\end{equation}
Under the maintained analytical DGP, the covariance terms cancel, yielding
\begin{equation}
 \operatorname{Cov}(\eta_j,\Delta_l)=0,
\qquad\forall j,l.   
\end{equation}
Because the vector is approximately jointly normal, zero covariance with every component of $(\Delta_1,\ldots,\Delta_J)$ implies
\begin{equation}
\eta_j\perp(\Delta_1,\ldots,\Delta_J),
\qquad\forall j.
\end{equation}
The same covariance calculation, together with independence between $X_1$ and the uninformative features, gives
\begin{equation}
\operatorname{Cov}(\Delta_1,\Delta_j)=0,
\qquad j>1.
\end{equation}
Joint normality then implies
\begin{equation}\label{eq:delta_independence}
\Delta_1\perp(\Delta_2,\ldots,\Delta_J).
\end{equation}

The algorithm selects the splitting feature \( \hat\ell \) to maximize the absolute difference between child node estimates:
\begin{equation}
\hat\ell = \arg\max_j |\Delta_j|. \label{eq:split_rule}
\end{equation}
With balanced binary features, this is analogous to the splitting rule in Equation~\eqref{eq:split_criterion}.

For each candidate feature \(k\), define the largest competing contrast as:
\begin{equation}
    M_{-k} = \max_{j \neq k} |\Delta_j|.
\end{equation}
Conditioning on \( \hat\ell = k \) implies that \( |\Delta_k| \) was the largest among all candidate features:
\begin{equation}\label{eq:max_selection}
    \hat\ell=k \Leftrightarrow |\Delta_k| > M_{-k}. 
\end{equation}
Because the subsequent derivations mainly concern the informative feature, we write
\begin{equation}
M:=M_{-1}=\max_{j>1}|\Delta_j|.
\end{equation}
By Equation~\eqref{eq:delta_independence}, $\Delta_1\perp M$.

The AE estimate is:
\begin{equation}
\hat\beta(\mathbf{x}) = \hat\tau_{\hat\ell}(x_{\hat\ell}). \label{eq:AE_estimate}
\end{equation}

Throughout the appendix, we focus on estimation for an individual \( \mathbf{x} = (1, x_2, \dots, x_J) \) with 
$\beta(\mathbf{x}) = \theta$.

\subsection{No Estimation Bias for Uninformative Splits} \label{app:adaptive_bias_uninformative}
\textbf{Proof for Equation~\eqref{eq:adaptive_bias_uninformative}},
\[
\mathbb{E}[\hat\beta(\mathbf{x}) \mid \hat\ell \ne 1] = 0.
\]

\textit{Proof.}
Fix $k>1$ and suppose $\hat\ell=k$. We first consider the case $x_k = 1$.
\begin{equation}
\mathbb{E}[\hat\beta(\mathbf{x})\mid\hat\ell=k]=\mathbb{E}[\hat\tau_k(1)\mid\hat\ell=k]
\end{equation}
Using the decomposition in Equation~\eqref{eq:tau1_decomp}, we get:
\begin{equation}
\mathbb{E}[\hat\tau_k(1)\mid\hat\ell=k]
=\frac{1}{2}\mathbb{E}[\eta_k\mid\hat\ell=k]
+\frac{1}{2}\mathbb{E}[\Delta_k\mid\hat\ell=k]
\end{equation}

Because the event $\{\hat\ell=k\}$ is a function of $(\Delta_1,\ldots,\Delta_J)$ alone and $\eta_k$ is independent of this vector,
\begin{equation}
\mathbb{E}[\eta_k\mid \hat\ell=k]
=\mathbb{E}[\eta_k]=0
\end{equation}

Let $\boldsymbol\Delta_{-1}=(\Delta_2,\ldots,\Delta_J).$ Because $\boldsymbol\Delta_{-1}$ is a centered Gaussian vector,
\begin{equation}
\boldsymbol\Delta_{-1}
\overset d=
-\boldsymbol\Delta_{-1},
\end{equation}
regardless of the correlations among its components. Moreover, because $\Delta_1\perp\boldsymbol\Delta_{-1}$,
\begin{equation}
(\Delta_1,\boldsymbol\Delta_{-1})
\overset d=
(\Delta_1,-\boldsymbol\Delta_{-1}).
\end{equation}
Recall that $\{\hat\ell=k\}=\{|\Delta_k|>M_{-k}\}$. Because this event depends only on the absolute values of the $\Delta_j$'s, replacing $\boldsymbol\Delta_{-1}$ with $-\boldsymbol\Delta_{-1}$ leaves the event unchanged while changing the sign of $\Delta_k$. By the distributional symmetry,
\begin{equation}
\mathbb E\!\left[\Delta_k\mathbf 1\{\hat\ell=k\}\right]
=
\mathbb E\!\left[-\Delta_k\mathbf 1\{\hat\ell=k\}\right]
=
-\mathbb E\!\left[\Delta_k\mathbf 1\{\hat\ell=k\}\right].
\end{equation}

Therefore,
\begin{equation}
\mathbb E\!\left[
\Delta_k\mathbf 1\{\hat\ell=k\}
\right]
=0.
\end{equation}
Dividing by $\mathbb P(\hat\ell=k)>0$ gives
\begin{equation}
\mathbb E[\Delta_k\mid\hat\ell=k]
=0.
\end{equation}
Combining the two terms gives $\mathbb E[\hat\beta(\mathbf x)\mid\hat\ell=k]=0$ when $x_k=1$. By the decomposition in Equation~\eqref{eq:tau0_decomp}, the same argument applies when $x_k = 0$. 

Thus, $\mathbb E[\hat\beta(\mathbf x)\mid\hat\ell=k]=0$ for either value of $x_k$ and every $k>1$, implying $\mathbb E[\hat\beta(\mathbf x)\mid\hat\ell\neq1]=0$.

\subsection{Upward Estimation Bias for Informative Splits} \label{app:adaptive_bias_informative}
\textbf{Proof for Equation~\eqref{eq:adaptive_bias_informative}},
\[
\mathbb{E}[\hat\beta(\mathbf{x}) \mid \hat\ell = 1] > \theta.
\]

\textit{Proof.}
If $\hat\ell=1$, the estimate for an individual with $x_1 = 1$ is $\hat\beta(\mathbf{x}) = \hat\tau_1(1)$. Using Equations~\eqref{eq:tau1_decomp} and~\eqref{eq:max_selection} we get:
\begin{align}
\mathbb{E}[\hat\beta(\mathbf{x}) \mid \hat\ell = 1]
&= \frac{1}{2} \mathbb{E}\left[\eta_1 + \Delta_1 \mid |\Delta_1| > M\right]
\end{align}

Because $\eta_1$ is independent of $(\Delta_1,\cdots,\Delta_J)$,
\begin{equation}
\mathbb{E}\left[\eta_1 \mid |\Delta_1| > M\right] = \mathbb{E}[\eta_1] = \mathbb{E}[\hat\tau_1(1)] + \mathbb{E}[\hat\tau_1(0)] = \theta + (-\theta) = 0,
\end{equation}
so:
\begin{equation}
\mathbb{E}[\hat\beta(\mathbf{x}) \mid \hat\ell = 1]
= \frac{1}{2} \mathbb{E}\left[\Delta_1 \mid |\Delta_1| >M\right].
\end{equation}

Let $\Delta_1 = 2\theta + \epsilon_1$, where $\epsilon_1 \sim \mathcal{N}(0, 2\sigma_1^2)$. Then:
\begin{equation}
\mathbb{E}[\Delta_1 \mid \hat\ell = 1]
= \mathbb{E}[2\theta + \epsilon_1 \mid |\Delta_1| > M]
= 2\theta + \mathbb{E}[\epsilon_1 \mid \epsilon_1 > M - 2\theta \ \text{or} \ \epsilon_1 < -M - 2\theta].
\end{equation}
To prove the inequality, it suffices to show:
\begin{equation}
\mathbb{E}[\epsilon_1 \mid \epsilon_1 > M - 2\theta \ \text{or} \ \epsilon_1 < -M - 2\theta] > 0.
\end{equation}

We can write:
\begin{equation}
\mathbb{E}[\epsilon_1 \mid \epsilon_1 > M - 2\theta \ \text{or} \ \epsilon_1 < -M - 2\theta] 
= \frac{\mathbb{E}[\epsilon_1 \cdot \mathbf{1}_{\epsilon_1 > M - 2\theta \ \text{or} \ \epsilon_1 < -M - 2\theta}]}{\mathbb{P}(\epsilon_1 > M - 2\theta \ \text{or} \ \epsilon_1 < -M - 2\theta)}.
\label{eq:conditional_expectation_fraction}
\end{equation}
The denominator is positive, so we analyze the numerator. Using the law of total expectation:
\begin{equation}
\mathbb{E}[\epsilon_1 \cdot \mathbf{1}_{\epsilon_1 > M - 2\theta \ \text{or} \ \epsilon_1 < -M - 2\theta}]
= \mathbb{E}_M \left[ \int_{M - 2\theta}^\infty t f(t) dt + \int_{-\infty}^{-M - 2\theta} t f(t) dt \right],
\label{eq:expectation_split_integral}
\end{equation}
where $f$ is the PDF of $\epsilon_1$. By the symmetry around 0 of the normal distribution, we have:
\begin{equation}
\int_{M - 2\theta}^\infty t f(t) dt + \int_{-\infty}^{-M - 2\theta} t f(t) dt
= \int_{|M - 2\theta|}^{M + 2\theta} t f(t) dt.
\label{eq:symmetric_integral}
\end{equation}
Therefore, the numerator in Equation~\eqref{eq:conditional_expectation_fraction} is positive, and we conclude:
\begin{equation}
\mathbb{E}[\Delta_1 \mid \hat\ell = 1] > 2\theta,
\quad \text{and so} \quad
\mathbb{E}[\hat\beta(\mathbf{x}) \mid \hat\ell = 1] > \theta.
\label{eq:adaptive_positive_bias}
\end{equation}

\subsection{AE Has Higher Probability of Selecting the Informative Feature}

\textbf{Proof for Equation~\eqref{eq:adaptive_probability}},
\[
\mathbb{P}(\hat\ell^\text{A} = 1) \ge \mathbb{P}(\hat\ell^\text{~H} = 1).
\]

Recall that
\[
\mathbb{P}(\hat\ell = 1) = \mathbb{P}(|\Delta_1| > M).
\]
We show that this probability increases with the sample size $n$, making AE more likely to identify the informative feature. This is because AE has lower standard errors $\sigma_j$ than HE, which uses only half the data for splitting.

\textit{Proof.}
Let $\lambda=\sqrt{n}$. Given that standard errors $\sigma_j$ scale with $1/\sqrt{n}$, we write:
\begin{equation}
\sqrt{2}\sigma_j=\frac{c_j}{\lambda},    
\end{equation}
for constants $c_j$. Then:
\begin{align}
    \Delta_1&=2\theta+\frac{c_1Z_1}{\lambda},\quad Z_1\sim\mathcal{N}(0,1),\\
    \Delta_j&=\frac{c_jZ_j}{\lambda}, \quad j>1, \quad Z_j\sim\mathcal{N}(0,1).
\end{align}
It follows that
\begin{equation}
M = \max_{j > 1} |\Delta_j| = \frac{1}{\lambda} \max_{j > 1} |c_jZ_j| = \frac{V}{\lambda}, \quad \text{where } V = \max_{j > 1} |c_jZ_j|.    
\end{equation}
Thus, the probability of selecting the informative feature is:
\begin{equation}
\mathbb{P}(\hat\ell = 1) = \mathbb{P}\left(\left|2\theta + \frac{c_1 Z_1}{\lambda}\right| > \frac{V}{\lambda}\right) = \mathbb{P}\left(|2\theta\lambda + c_1 Z_1| > V\right).
\end{equation}
Taking the expectation over \( V \), we get:
\begin{equation}
\mathbb{P}(\hat\ell = 1) = \mathbb{E}_V\left[\mathbb{P}(|2\theta\lambda + c_1 Z_1| > V \mid V)\right].
\end{equation}
It suffices to show that for each \( v > 0 \), the conditional probability \( \mathbb{P}(|2\theta\lambda + c_1 Z_1| > v) \) increases with \( \lambda \). Define this probability as $g(\lambda)$:
\begin{align}
g(\lambda) &= \mathbb{P}(|2\theta\lambda+c_1Z_1|>v)\\ 
&= \mathbb{P}(2\theta\lambda+c_1Z_1>v) + \mathbb{P}(2\theta\lambda+c_1Z_1<-v)\\
&= 1 - \Phi\left(\frac{v-2\theta\lambda}{c_1}\right) + \Phi\left(\frac{-v-2\theta\lambda}{c_1}\right) 
\end{align}
and compute its derivative:
\begin{align}
g'(\lambda) &= \frac{2\theta}{c_1}\cdot\left[
\phi\left(\frac{v-2\theta\lambda}{c_1}\right)-\phi\left(\frac{-v-2\theta\lambda}{c_1}\right)\right] \\
&= \frac{2\theta}{c_1}\cdot\left[
\phi\left(\frac{v-2\theta\lambda}{c_1}\right)-\phi\left(\frac{v+2\theta\lambda}{c_1}\right)\right] 
\end{align}
where \( \phi \) is the standard normal PDF. Because \( \phi \) is symmetric and strictly decreasing on \( (0, \infty) \),  for any $v>0$, we have:
\begin{equation}
\left|\frac{v-2\theta\lambda}{c_1}\right|<\frac{v+2\theta\lambda}{c_1},\quad 
\text{so } \phi\left(\frac{v-2\theta\lambda}{c_1}\right)>\phi\left(\frac{v+2\theta\lambda}{c_1}\right)
\end{equation}
so \( g'(\lambda) > 0 \). Thus, \( \mathbb{P}(|2\theta\lambda + c_1 Z_1| > v) \) increases with \( \lambda \), and therefore \( \mathbb{P}(\hat\ell = 1) \) increases with \( n \).

\subsection{Adaptive Estimation Has Lower Bias}

\textbf{Proof for Equation~\eqref{eq:adaptive_lower_bias}},
\[
\label{eq:adaptive_lower_bias_restated}
\left| \mathbb{E}[\hat{\beta}^{\text{A}}(\mathbf{x})] - \beta(\mathbf{x}) \right|
\leq
\left| \mathbb{E}[\hat{\beta}^{\text{H}}(\mathbf{x})] - \beta(\mathbf{x}) \right|.
\]

We already established that
\begin{equation*}
\label{eq:adaptive_expectation_larger}
\mathbb{E}[\hat{\beta}^{\text{A}}]
>
\mathbb{E}[\hat{\beta}^{\text{H}}]
\end{equation*}
because AE overestimates the SPATE when it splits on the informative feature and is unbiased when it splits on a non-informative feature. It therefore suffices to show that
\begin{equation}
\label{eq:adaptive_expectation_below_theta}
\mathbb{E}[\hat\beta^{\text{A}}(\mathbf{x})] < \theta,
\end{equation}
which we establish next.

\textit{Proof.} To simplify the notation, let $\hat\beta=\hat\beta^{\text{A}}$. By the law of total expectation,
\begin{equation}
\label{eq:law_total_expectation_adaptive}
\mathbb{E}[\hat\beta(\mathbf{x})]
=
\mathbb{E}[\hat\beta(\mathbf{x}) \mid \hat\ell = 1] \cdot \mathbb{P}(\hat\ell = 1)
+
\mathbb{E}[\hat\beta(\mathbf{x}) \mid \hat\ell \ne 1] \cdot \mathbb{P}(\hat\ell \ne 1).
\end{equation}

From previous results (Appendices~\ref{app:adaptive_bias_uninformative} and~\ref{app:adaptive_bias_informative}),
\begin{itemize}
  \item \( \mathbb{E}[\hat\beta(\mathbf{x}) \mid \hat\ell \ne 1] = 0 \),
  \item \( \mathbb{E}[\hat\beta(\mathbf{x}) \mid \hat\ell = 1]
  = \frac{1}{2} \mathbb{E}[\Delta_1 \mid |\Delta_1| > M] \),
\end{itemize}
so
\begin{equation}
\label{eq:adaptive_expectation_expanded}
\mathbb{E}[\hat\beta(\mathbf{x})]
=
\frac{1}{2}
\cdot
\mathbb{P}(|\Delta_1| > M)
\cdot
\mathbb{E}[\Delta_1 \mid |\Delta_1| > M].
\end{equation}

Now consider the decomposition of the unconditional expectation:
\begin{equation}
\label{eq:theta_decomposition}
\theta
=
\frac{1}{2} \mathbb{E}[\Delta_1]
=
\frac{1}{2}
\left(
\mathbb{P}(|\Delta_1| > M) \cdot \mathbb{E}[\Delta_1 \mid |\Delta_1| > M]
+
\mathbb{P}(|\Delta_1| \le M) \cdot \mathbb{E}[\Delta_1 \mid |\Delta_1| \le M]
\right).
\end{equation}

Subtracting Equation \eqref{eq:adaptive_expectation_expanded} from Equation \eqref{eq:theta_decomposition} gives
\begin{equation}
\label{eq:theta_minus_adaptive}
\theta - \mathbb{E}[\hat\beta(\mathbf{x})]
=
\frac{1}{2}
\cdot
\mathbb{P}(|\Delta_1| \le M)
\cdot
\mathbb{E}[\Delta_1 \mid |\Delta_1| \le M].
\end{equation}

We now show that $\mathbb{E}[\Delta_1 \mid |\Delta_1| \le M] > 0$. Because \( \Delta_1 \sim \mathcal{N}(2\theta, 2\sigma_1^2) \), the density \( f_{\Delta_1}(t) \) is symmetric around \( 2\theta > 0 \). For any \( t > 0 \),
\begin{equation}
\label{eq:density_asymmetry}
|t - 2\theta| < |-t - 2\theta|
\quad \Rightarrow \quad
f_{\Delta_1}(t) > f_{\Delta_1}(-t).
\end{equation}

Thus,
\begin{align}
\label{eq:truncated_expectation_integral}
\mathbb{E}[\Delta_1 \mid |\Delta_1| \le z]
&=
\frac{1}{\mathbb{P}(|\Delta_1| \le z)}
\int_{-z}^{z} t f_{\Delta_1}(t)\, dt \nonumber\\
&=
\frac{1}{\mathbb{P}(|\Delta_1| \le z)}
\int_{0}^{z}
t \big[ f_{\Delta_1}(t) - f_{\Delta_1}(-t) \big]\, dt
> 0.
\end{align}

This holds pointwise for every \( z \ge 0 \), so 
\begin{equation}
\label{eq:truncated_expectation_M}
\mathbb{E}[\Delta_1 \mid |\Delta_1| \le M] > 0,
\end{equation}
and therefore $\mathbb{E}[\hat\beta(\mathbf{x})] < \theta$ holds.

\section{Analytical Derivations for Variance Analysis}\label{app:variance}

This appendix provides analytical derivations that clarify how honesty impacts each of the sources of tree--tree dependence introduced in Equation~\eqref{eq:covariance-decomposition} of Section~\ref{sec:variance},
\[
\mathrm{Cov}\!\big(\hat\beta_b^{\text{tree}},\hat\beta_{b'}^{\text{tree}}\big)
=
\underbrace{\mathrm{Cov}\!\big(\tau_b,\tau_{b'}\big)}_{\text{Target coupling}}
+
\underbrace{\mathrm{Cov}\!\big(\varepsilon_b,\varepsilon_{b'}\big)}_{\text{Noise overlap}}
+
\underbrace{\mathrm{Cov}\!\big(\tau_b,\varepsilon_{b'}\big)
+
\mathrm{Cov}\!\big(\varepsilon_b,\tau_{b'}\big)}_{\text{Target--estimation spillover}}.
\]

\subsection{Implications of Honesty for Target Coupling}
\label{app:approx-variance}

Honesty affects target coupling through its impact on split selection. Target coupling is the covariance between the approximation targets of two trees evaluated at the same individual $\mathbf{x}$,
\begin{equation*}
\label{eq:target-coupling-def}
\mathrm{Cov}(\tau_b,\tau_{b'}),
\qquad
\tau_b
=
\tau\!\left(\hat\ell(\mathbf{x};\mathcal{S}^{(b)}_{\text{sp}})\right),
\end{equation*}
where randomness arises from sampling variation in the splitting samples $\mathcal{S}^{(b)}_{\text{sp}}$. 

Because $\tau_b$ and $\tau_{b'}$ are identically distributed, target coupling can be expressed as
\begin{equation}
\label{eq:target-coupling-var-corr}
\mathrm{Cov}(\tau_b,\tau_{b'})
=
\mathrm{Corr}(\tau_b,\tau_{b'}) \cdot \mathrm{Var}(\tau_b).
\end{equation}

Honesty affects target coupling by reducing the size of the splitting sample, denoted by
\begin{equation}
m = \lvert \mathcal{S}^{(b)}_{\text{sp}} \rvert .    
\end{equation}
Differentiating Equation~\eqref{eq:target-coupling-var-corr} with respect to $m$ yields
\begin{equation}
\label{eq:cov-derivative}
\frac{\partial}{\partial m}\mathrm{Cov}(\tau_b,\tau_{b'})
=
\mathrm{Corr}(\tau_b,\tau_{b'})\,\frac{\partial}{\partial m}\mathrm{Var}(\tau_b)
+
\mathrm{Var}(\tau_b)\,\frac{\partial}{\partial m}\mathrm{Corr}(\tau_b,\tau_{b'}).
\end{equation}

We focus on the first term. The correlation between $\tau_b$ and $\tau_{b'}$ arises from the fact that trees are grown on overlapping subsamples of the same training sample, which induces shared successes and failures in split selection. While the degree of overlap determines the level of this dependence, the size of the splitting sample primarily affects the dispersion of approximation targets across trees. Accordingly, we assume that either (i) the derivatives of the variance and correlation terms with respect to $m$ have the same sign, or (ii) the effect of $m$ on the correlation term is not large enough to overturn the sign of the variance term. Under this assumption, the sign of $\partial\,\mathrm{Cov}(\tau_b,\tau_{b'})/\partial m$ is governed by the behavior of $\partial\,\mathrm{Var}(\tau_b)/\partial m$. The remainder of this appendix therefore analyzes $\mathrm{Var}(\tau_b)$ as a function of the splitting-sample size.

\vspace{0.5em}
\noindent
Fix an individual $\mathbf{x}$ with CATE $\beta$ that may be assigned to candidate leaves $\ell_1,\ldots,\ell_J$, each associated with a SPATE $\tau_j = \tau(\ell_j)$. Let
\begin{equation}
\label{eq:pj-def}
p_j(m) = \Pr(\hat\ell = \ell_j)
\end{equation}
denote the probability that $\mathbf{x}$ is assigned to leaf $\ell_j$ when the splitting sample has size $m$. Among the candidate leaves, define the best available approximation target as
\begin{equation}
\label{eq:tau-star}
\tau^\star = \arg\min_{\tau_j} |\tau_j-\beta|.
\end{equation}
Define the mean approximation target
\begin{equation}
\label{eq:mu-def}
\mu(m) := \mathbb{E}[\tau_b] = \sum_{j=1}^J p_j(m)\tau_j,
\end{equation}
and the variance of approximation targets
\begin{equation}
\label{eq:V-def}
V(m) := \mathrm{Var}(\tau_b)
= \sum_{j=1}^J p_j(m)\big(\tau_j - \mu(m)\big)^2.
\end{equation}
Differentiating with respect to $m$ yields
\begin{equation}
\label{eq:Vprime-general}
V'(m)
=
\sum_{j=1}^J p'_j(m)\,(\tau_j-\mu(m))^2,
\end{equation}
where $\sum_{j=1}^J p_j(m)=1$ implies $\sum_{j=1}^J p'_j(m)=0$. Thus, changes in $V(m)$ arise entirely from reallocation of probability mass across candidate approximation targets. Equation~\eqref{eq:Vprime-general} shows that the sign of $V'(m)$ depends on where probability mass shifts, weighted by squared deviations from the current mean target $\mu(m)$.

To interpret Equation~\eqref{eq:Vprime-general}, we define the \emph{correctable approximation bias} as the gap between the expected approximation target and the best available target, $|\mu(m)-\tau^\star|$. The effect of the splitting-sample size $m$ on the variance of the approximation target—and, through it, on target coupling—depends on the magnitude of this bias. When correctable approximation bias is large, increasing $m$ raises the variance of the approximation target; when it is small, increasing $m$ reduces that variance. Because honesty reduces the effective splitting-sample size, it therefore dampens target coupling when correctable approximation bias is large and amplifies it when the bias is small. We now explain the mechanism underlying this result.

When correctable approximation bias is large, split selection frequently assigns $\mathbf{x}$ to suboptimal leaves even though substantially better targets exist. Increasing the splitting-sample size improves split selection and shifts probability mass toward those better targets. Because $\mu(m)$ lies near suboptimal values in this regime, the targets receiving additional probability mass are far from the current mean. As a result, the squared deviations $(\tau_j-\mu(m))^2$ are large precisely for those $j$ with $p'_j(m)>0$. Equation~\eqref{eq:Vprime-general} therefore implies that increasing $m$ can increase $\mathrm{Var}(\tau_b)$: approximation bias falls as split selection improves, but approximation variance rises because probability mass shifts toward better yet more extreme targets.

In contrast, when correctable approximation bias is small, the expected approximation target $\mu(m)$ is already close to $\tau^\star$. Further improvements in split selection then primarily reallocate probability mass among targets that are already close to $\mu(m)$. This may occur either because split selection is highly reliable, or because target selection is intrinsically uninformative for $\mathbf{x}$—for example, when the features provide little leverage to predict effect heterogeneity or when the individual effect lies near the population average. In either case, the targets receiving additional probability mass have similar values to $\mu(m)$, so the squared deviations $(\tau_j-\mu(m))^2$ are small. Equation~\eqref{eq:Vprime-general} therefore implies that increasing $m$ reduces $\mathrm{Var}(\tau_b)$ through further concentration.

As a final example, we return to the stylized setting Section~\ref{sec:example}, where $\tau^\star=\tau_1=\theta$ and $\tau_j=0$ for all $j>1$. In this case, split selection reduces to a binary choice between $\tau^\star$ and a common alternative. When $p_1(m)<1/2$, increasing $m$ shifts probability mass toward $\tau^\star=\theta$, which lies far from the current mean $\mu(m)=p_1(m)\theta$, so Equation~\eqref{eq:Vprime-general} implies that $V(m)$ increases. Once $p_1(m)>1/2$, the mean approximation target lies closer to $\tau^\star$, and further increases in $m$ primarily concentrate probability mass on that same target, causing $V(m)$ to decrease. This behavior explains Figure~\ref{fig:target-coupling}.

\subsection{Implications of Honesty for Target--Estimation Spillover}

We analyze next the target--estimation spillover terms
$\mathrm{Cov}\!\big(\tau_b,\varepsilon_{b'}\big)$ and 
$\mathrm{Cov}\!\big(\varepsilon_b,\tau_{b'}\big)$, which capture dependence between the approximation target of one tree and the estimation error of another. We show that both terms equal zero under forest honesty. Forest honesty implies that
\begin{equation*}
\mathbb{E}[\varepsilon_b \mid \hat\ell_b]=0
\quad\text{and}\quad
\mathbb{E}[\varepsilon_b \mid \hat\ell_b,\hat\ell_{b'}]=0.
\label{eq:forest-honesty-app}
\end{equation*}
We have that
\begin{align}
\mathrm{Cov}\!\big(\varepsilon_{b}, \tau_{b'}\big)
&=
\mathbb{E}\!\left[\tau_{b'}\,\varepsilon_{b}\right]
-
\mathbb{E}[\tau_{b'}]\mathbb{E}[\varepsilon_{b}].
\label{eq:cov-start}
\end{align}
Applying the law of iterated expectations, 
\begin{equation}
 \mathbb{E}[\varepsilon_{b}]=\mathbb{E}\left[\mathbb{E}[\varepsilon_{b} \mid \hat\ell_b]\right]=0,  
\end{equation}
so it suffices to show that $\mathbb{E}[\tau_{b'}\,\varepsilon_{b}]=0$. If we apply the law of iterated expectations again,
\begin{align}
\mathbb{E}\!\left[\tau_{b'}\,\varepsilon_{b}\right]
&=
\mathbb{E}\!\left[
\tau_{b'}\,
\mathbb{E}\!\left[\varepsilon_{b} \mid \hat\ell_b,\hat\ell_{b'}\right]
\right]=0,
\label{eq:iter-exp}
\end{align}
because $\tau_{b'}$ is deterministic given $\hat\ell_{b'}$. Therefore
\begin{equation}
\mathrm{Cov}\!\big(\varepsilon_b,\tau_{b'}\big)=0.
\label{eq:cov-zero-1}
\end{equation}
An identical argument yields
\begin{equation}
\mathrm{Cov}\!\big(\tau_b,\varepsilon_{b'}\big)=0.
\label{eq:cov-zero-2}
\end{equation}

\subsection{Implications of Honesty for Noise Overlap}

Finally, we analyze how honesty affects noise overlap, decomposed as
\begin{equation*}
\mathrm{Cov}(\varepsilon_b,\varepsilon_{b'})
=
\mathbb{E}\!\left[
\mathrm{Cov}(\varepsilon_b,\varepsilon_{b'} \mid \hat\ell_b, \hat\ell_{b'})
\right]
+
\mathrm{Cov}\!\left(
\mathbb{E}[\varepsilon_b \mid \hat\ell_b, \hat\ell_{b'}],
\mathbb{E}[\varepsilon_{b'} \mid \hat\ell_b, \hat\ell_{b'}]
\right).
\label{eq:noise-overlap-decomposition}
\end{equation*}

Under forest honesty, the second term vanishes because
$\mathbb{E}[\varepsilon_b \mid \hat\ell_b, \hat\ell_{b'}]=0$ for all $b \neq b'$.
This appendix therefore focuses on the remaining term.

Recall from Equation~\eqref{eq:tree-variance-decomp} that each tree prediction can be written as
\begin{equation*}
\hat\beta_b^{\text{tree}} = \tau_b + \varepsilon_b,
\qquad
\varepsilon_b = \hat\tau_b - \tau_b,
\end{equation*}
where $\tau_b=\tau(\hat\ell_b)$ is the SPATE associated with leaf $\hat\ell_b$ and
$\hat\tau_b$ is its estimate based on the estimation subsample
$\mathcal S^{(b)}_{\text{es}}$.
Because $\tau_b$ and $\tau_{b'}$ are fixed given $(\hat\ell_b,\hat\ell_{b'})$,
\begin{equation}
\mathrm{Cov}(\varepsilon_b,\varepsilon_{b'} \mid \hat\ell_b,\hat\ell_{b'})
=
\mathrm{Cov}(\hat\tau_b,\hat\tau_{b'} \mid \hat\ell_b,\hat\ell_{b'}).
\end{equation}

For analytical convenience, we assume SPATEs are estimated as sample means of unbiased, noisy signals of the CATE:
\begin{equation}
\hat\tau(\ell;\mathcal S_{\text{es}})
=
\frac{1}{|\mathcal I(\ell;\mathcal S_{\text{es}})|}
\sum_{i\in\mathcal I(\ell;\mathcal S_{\text{es}})} Z_i,
\label{eq:spate-mean-estimator}
\end{equation}
where $\mathcal I(\ell;\mathcal S_{\text{es}})$ denotes the set of estimation-sample observations
falling in leaf $\ell$.
In a randomized experiment, one convenient choice for the signal is the transformed outcome
\begin{equation}
Z
=
\frac{T Y}{\omega(\mathbf X)}
-
\frac{(1-T)Y}{1-\omega(\mathbf X)},
\end{equation}
which satisfies $\mathbb{E}[Z \mid \mathbf X] = \beta(\mathbf X)$.
The specific form of $Z$ is not important; all arguments below rely only on
$Z_i$ being i.i.d.\ with mean $\tau(\ell)$ within each leaf.

Under this representation, the estimator for tree $b$ is
\begin{equation}
\hat\tau_b
=
\frac{1}{n_b}
\sum_{i\in\mathcal I_b} Z_i,
\qquad
\mathcal I_b=\mathcal I(\hat\ell_b;\mathcal S^{(b)}_{\text{es}}),
\end{equation}
with $n_b = |\mathcal I_b|$. In addition, let
\begin{equation}
p_b = \mathbb{P}(X_i \in \hat\ell_b),
\quad
p_{b'} = \mathbb{P}(X_i \in \hat\ell_{b'}),
\quad
p_{bb'} = \mathbb{P}(X_i \in \hat\ell_b \cap \hat\ell_{b'}).
\end{equation}
The training sample contains $N$ observations, and for each tree $b$ the estimation sample $\mathcal S^{(b)}_{\text{es}}$ is drawn by subsampling at rate $p_{\text{es}}$, independently of leaf assignment due to honesty.

To evaluate the covariance of the SPATE estimators, we must account for two sources of randomness:
(i) noise in the signals $\{Z_i\}$ and
(ii) random estimation subsamples $(\mathcal I_b,\mathcal I_{b'})$.
We therefore apply the law of total covariance over the subsampling mechanism:
\begin{align}
\mathrm{Cov}(\hat\tau_b,\hat\tau_{b'}\mid \hat\ell_b,\hat\ell_{b'})
&=
\mathbb{E}\!\Big[
\mathrm{Cov}(\hat\tau_b,\hat\tau_{b'}\mid \hat\ell_b,\hat\ell_{b'},\mathcal I_b,\mathcal I_{b'})\mid \hat\ell_b,\hat\ell_{b'}
\Big] \notag\\
&\quad+
\mathrm{Cov}\!\Big(
\mathbb{E}[\hat\tau_b\mid \hat\ell_b,\hat\ell_{b'},\mathcal I_b,\mathcal I_{b'}],
\mathbb{E}[\hat\tau_{b'}\mid \hat\ell_b,\hat\ell_{b'},\mathcal I_b,\mathcal I_{b'}]\mid \hat\ell_b,\hat\ell_{b'}
\Big).
\label{eq:totcov-subsampling}
\end{align}

Because estimation subsamples are drawn uniformly at random under honesty, inclusion into the estimation sample is independent of the outcome signal $Z$, so
\begin{equation}
\mathbb{E}[\hat\tau_b\mid \hat\ell_b,\hat\ell_{b'},\mathcal I_b,\mathcal I_{b'}]=\tau_b,
\qquad
\mathbb{E}[\hat\tau_{b'}\mid \hat\ell_b,\hat\ell_{b'},\mathcal I_b,\mathcal I_{b'}]=\tau_{b'}.
\end{equation}
Hence the second term in Equation~\eqref{eq:totcov-subsampling} is zero. Using the sample-mean representation
of $\hat\tau_b$,
\begin{align}
\mathrm{Cov}(\hat\tau_b,\hat\tau_{b'}\mid \hat\ell_b,\hat\ell_{b'},\mathcal I_b,\mathcal I_{b'})
&=
\frac{1}{n_b n_{b'}}
\sum_{i\in\mathcal I_b}\sum_{j\in\mathcal I_{b'}}
\mathrm{Cov}(Z_i,Z_j \mid \hat\ell_b,\hat\ell_{b'},\mathcal I_b,\mathcal I_{b'}) \notag\\
&=
\frac{1}{n_b n_{b'}}
\sum_{i\in\mathcal I_b\cap \mathcal I_{b'}}
\mathrm{Var}(Z_i \mid \hat\ell_b,\hat\ell_{b'},\mathcal I_b,\mathcal I_{b'}) \notag\\
&=
\frac{1}{n_b n_{b'}}
\sum_{i\in\mathcal I_b\cap \mathcal I_{b'}}
\mathrm{Var}(Z_i \mid X_i \in \hat\ell_b\cap \hat\ell_{b'}) \notag\\
&=
\frac{\sigma^2_{bb'}\, m_{bb'}}{n_b n_{b'}},
\label{eq:cov-given-subsamples}
\end{align}
where $m_{bb'}=|\mathcal I_b\cap\mathcal I_{b'}|$ and
$\sigma^2_{bb'}=\mathrm{Var}(Z_i\mid X_i\in\hat\ell_b\cap\hat\ell_{b'})$.
The second line uses the i.i.d.\ assumption, which implies
$\mathrm{Cov}(Z_i,Z_j\mid \hat\ell_b,\hat\ell_{b'},\mathcal I_b,\mathcal I_{b'})=0$ for $i\neq j$, so only observations shared across both
estimation subsamples contribute to the covariance.
The third line follows from independence between the outcome signal and inclusion into the estimation subsample.

An observation contributes to $m_{bb'}$ only if it belongs to both leaf populations and is selected into both estimation subsamples. Conditional on $(\hat\ell_b,\hat\ell_{b'})$, the counts $(m_{bb'}, n_b, n_{b'})$ are sums of $N$ i.i.d.\ indicators with means
\begin{equation}
\mathbb{E}[m_{bb'}\mid \hat\ell_b,\hat\ell_{b'}] = p_{bb'}\,p_{\text{es}}^2\,N,
\quad
\mathbb{E}[n_b\mid \hat\ell_b]=p_b\,p_{\text{es}}\,N,
\quad
\mathbb{E}[n_{b'}\mid \hat\ell_{b'}]=p_{b'}\,p_{\text{es}}\,N.
\end{equation}
Applying the delta method to the smooth map $(x,y,z)\mapsto x/(yz)$ at these means, and using that the empirical proportions $m_{bb'}/N$, $n_b/N$, $n_{b'}/N$ fluctuate around their means at rate $O(N^{-1/2})$ under standard leaf-regularity conditions, gives
\begin{equation}
\mathbb{E}\!\left[\frac{m_{bb'}}{n_b n_{b'}} \,\middle|\, \hat\ell_b,\hat\ell_{b'}\right]
= \frac{p_{bb'}}{p_b p_{b'}\,N} + O(N^{-2}),
\end{equation}
where the subsampling rate $p_{\text{es}}$ cancels between numerator and denominator. Substituting into Equation~\eqref{eq:cov-given-subsamples},
\begin{equation}
\mathrm{Cov}(\hat\tau_b,\hat\tau_{b'}\mid \hat\ell_b,\hat\ell_{b'})
= \sigma^2_{bb'}\,\frac{p_{bb'}}{p_b p_{b'}\,N} + O(N^{-2}).
\label{eq:cov-geometry}
\end{equation}

Equation~\eqref{eq:cov-geometry} shows that, asymptotically, the covariance depends only on the geometric overlap between the two leaf populations. The cancellation of $p_{\text{es}}$ reflects that honesty's reduction of estimation sample sizes is exactly offset by the corresponding reduction in shared observations, so honesty does not affect asymptotic noise overlap.

\section{AE--HE Selection Rule}\label{app:selection_rule}

This appendix motivates and analyzes the rule we use to select between AE and HE when both pass the heterogeneity test. We first explain why the selection problem calls for an asymmetric rule that favors HE (\ref{app:sel_logic}), then formalize the rule (\ref{app:sel_rule}), discuss the choice of threshold (\ref{app:sel_threshold}), and present a sensitivity analysis (\ref{app:sel_sensitivity}). A final subsection discusses the implementation of the heterogeneity test used as a screening step (\ref{app:sel_test}).

\subsection{Why an Asymmetric Rule}\label{app:sel_logic}

Our theoretical analysis implies that the selection between AE and HE is a choice between a flexible method and its regularized counterpart. That means which one is better depends on the SNR: AE tends to win when SNR is high, HE when it is low. Since the SNR is unknown, a natural strategy is to compare the two methods empirically and pick whichever performs better on held-out data.

The key observation is that the reliability of this empirical comparison also depends on the SNR. Cross-validation estimates based on transformed outcomes are reliable when the SNR is high, but not so much when it is low. So when data can reliably adjudicate the choice (high SNR), it points to AE; when it cannot (low SNR), theory points to HE.

This suggests an asymmetric rule that gets the best of both worlds: treat HE as the default, and switch to AE only when the empirical evidence in its favor is strong relative to the uncertainty in the comparison. When the signal is strong, AE's advantage is large and easily clears the evidence bar; when the signal is weak, the bar is harder to clear and the selector falls back on the method that theory favors in that regime. Relying on a noisy comparison alone—selecting whichever method posts the lower estimated error—would forgo this protection, in the same way that cross-validation can overfit when uncertainty in the validation estimates is ignored~\citep{lei2020cross}.

\subsection{The Decision Rule}\label{app:sel_rule}

We quantify the uncertainty in the AE--HE comparison using the standard error of the paired error difference. Because both models are evaluated on the same held-out data points, their errors are correlated; taking paired differences accounts for this shared variation and directly targets the uncertainty in the performance contrast. Paired differences are a standard tool in model comparison: \cite{dietterich1998approximate} compares learning algorithms using paired differences in prediction errors across train/test splits, and \cite{lei2020cross} uses the same object to construct confidence sets of competitive models that account for the overfitting induced by CV-based selection.

Formally, let $\bar{\Delta}$ denote the mean paired error difference, defined as HE minus AE, and let $\mathrm{SE}(\bar{\Delta})$ denote its standard error. The rule is:
\begin{equation}
\text{select AE if }
\bar{\Delta}
>
c\,\mathrm{SE}(\bar{\Delta});
\qquad
\text{otherwise, select HE},
\end{equation}
where $c \ge 0$ is a prespecified threshold governing how much evidence is required before the default is abandoned. At $c=0$, the rule reduces to selecting the model with the lower estimated MSE. As $c$ increases, AE must show a larger advantage before HE is abandoned, so the rule favors HE more often; as $c \to \infty$, the rule selects HE almost always.

\subsection{Choosing the Threshold}\label{app:sel_threshold}

The threshold $c$ formalizes what ``sufficiently strong evidence'' means, and there is no principled, universally agreed-upon value---just as there is no first-principles justification for the conventional 5\% significance level in hypothesis testing. What exist instead are conventions, and two are relevant here.

The first comes from framing the comparison as a hypothesis test. \cite{dietterich1998approximate} compares learning algorithms this way, using a conventional significance level such as $\alpha = 0.05$ to decide whether to reject the null of no difference. In our setting, a one-sided 5\% level corresponds approximately to $c = 1.6$. This threshold demands very strong evidence before selecting AE, which protects against costly false AE selections but sacrifices a large share of the available gains whenever AE is in fact the better choice---as it often is in the ACIC datasets. We quantify this cost in Section~\ref{app:sel_sensitivity}.

The second convention comes from the model selection literature: the one-standard-error (1-SE) rule of \cite{breiman1984classification}. In its conventional form, the rule selects the simplest (most regularized) model whose estimated error is within one standard error of the best model's error~\citep{hastie2009elements}. This principle has been widely used to select regularization parameters~\citep{breiman1984classification, hastie2009elements}, is the default behavior offered by the R package \texttt{glmnet}~\citep{friedman2010regularization}, and has been adopted across domains: \cite{villanueva2024system} use it to select parsimonious Lasso models for RNA and protein localization, and \cite{windmeijer2019use} apply it in Mendelian randomization to identify invalid instrumental variables. Applying the 1-SE threshold to paired performance differences also has precedent: \cite{hammond2024identifying} use it to compare eclipse-mapping models in astronomy, and \cite{piironen2020projective} apply the same logic in projection predictive model selection. Our rule is this heuristic applied to the AE--HE choice, with HE playing the role of the more regularized model: setting $c = 1$ selects HE whenever its estimated error is within one standard error of AE's.

We adopt $c=1$ as our default threshold, without claiming it is optimal.  The optimal threshold depends on quantities that are not known in practice, such as the relative cost and probability of the two possible mistakes---overfitting with AE versus underfitting with HE. A different threshold could reasonably be chosen instead. What matters for our purposes is that (i) the rule embeds the correct qualitative asymmetry, and (ii) its performance is robust across thresholds, as we show next in Section~\ref{app:sel_sensitivity}.

\subsection{Sensitivity Analysis}\label{app:sel_sensitivity}

Figure~\ref{fig:selection_threshold} presents a sensitivity analysis over $c \in [0, 1.6]$ using the ACIC datasets. As $c$ increases, the selector favors HE more often, and both mean regret and the error rate increase; this reflects that AE is the better choice in most of these datasets. The empirical selector outperforms both pure strategies in mean regret for all thresholds up to roughly $c = 1.2$; beyond that point, it falls behind defaulting to AE. Even so, it outperforms defaulting to HE across the entire range---and since defaulting to HE is the prevailing practice, the choice of $c$ affects how much of AE's potential gains are captured, not whether the procedure improves on the status quo.

\begin{figure}[t]
    \centering
    \pgfplotsset{
        thresholdstyle/.style={
            width=\linewidth,
            height=5.8cm,
            grid=major,
            grid style={dashed, gray!30},
            xmin=-0.05, xmax=1.65,
            xtick={0, 0.2, 0.4, 0.6, 0.8, 1.0, 1.2, 1.4, 1.6},
            xticklabels={0, 0.2, 0.4, 0.6, 0.8, 1.0, 1.2, 1.4, 1.6},
            xlabel={Threshold $c$},
            tick label style={font=\scriptsize},
            label style={font=\footnotesize},
            title style={font=\footnotesize},
        }
    }
    \begin{subfigure}[c]{0.365\textwidth}
        \begin{tikzpicture}
        \begin{axis}[
            thresholdstyle,
            title={$\mathbb{E}[\text{Regret}]$},
            ymin=0.01, ymax=0.036,
            ytick={0.010, 0.015, 0.020, 0.025, 0.030, 0.035},
            yticklabels={1.0, 1.5, 2.0, 2.5, 3.0, 3.5},
            scaled y ticks=base 10:2,
        ]
        \addplot[blue, thick, dashed] coordinates {(0, 0.01365) (1.6, 0.01365)}; 
        \addplot[red, thick, dashdotted] coordinates {(0,0.03437) (1.6,0.03437)}; 
        \addplot[
            mark=*, thick, solid, myempirical,
            mark size=1.5pt,
            mark indices={1,3,5,7,9,11,13,15,17}
        ] coordinates {
            (0.0, 0.01106) (0.1, 0.01092) (0.2, 0.01087) (0.3, 0.01091)
            (0.4, 0.01113) (0.5, 0.01128) (0.6, 0.01141) (0.7, 0.01141)
            (0.8, 0.01173) (0.9, 0.01204) (1.0, 0.01243) (1.1, 0.01274)
            (1.2, 0.01346) (1.3, 0.01415) (1.4, 0.0146) (1.5, 0.01566)
            (1.6, 0.0162)
        }; 
        \end{axis}
        \end{tikzpicture}
    \end{subfigure}
    \hspace{-0.06\textwidth}
    \begin{subfigure}[c]{0.365\textwidth}
        \begin{tikzpicture}
        \begin{axis}[
            thresholdstyle,
            title={Error rate (\%)},
            ymin=20, ymax=77,
            ytick={20, 30, 40, 50, 60, 70},
            yticklabels={20, 30, 40, 50, 60, 70}
        ]
        \addplot[blue, thick, dashed] coordinates {(0, 25.2) (1.6, 25.2)}; 
        \addplot[red, thick, dashdotted] coordinates {(0, 74.8) (1.6, 74.8)};   
        \addplot[
            mark=*, thick, solid, myempirical,
            mark size=1.5pt,
            mark indices={1,3,5,7,9,11,13,15,17}
        ] coordinates {
            (0.0, 23.67219) (0.1, 23.88635) (0.2, 24.25757) (0.3, 24.78584)
            (0.4, 25.24272) (0.5, 25.8281) (0.6, 26.55625) (0.7, 27.32724)
            (0.8, 28.59794) (0.9, 29.55454) (1.0, 30.71102) (1.1, 32.02456)
            (1.2, 33.63792) (1.3, 35.1799) (1.4, 36.69332) (1.5, 38.8064)
            (1.6, 40.39121)
        };     
        \end{axis}
        \end{tikzpicture}
    \end{subfigure}
    \hspace{-0.06\textwidth}
    \begin{subfigure}[c]{0.365\textwidth}
        \begin{tikzpicture}
        \begin{axis}[
            thresholdstyle,
            title={$\mathbb{E}[\text{Regret}\mid\text{error}]$},
            ymin=0.039, ymax=0.055,
            ytick={0.040,0.042,0.044,0.046,0.048,0.050,0.052,0.054},
            yticklabels={4.0,4.2,4.4,4.6,4.8,5.0,5.2,5.4},
            scaled y ticks=base 10:2,
        ]
        \addplot[blue, thick, dashed] coordinates {(0, 0.05416) (1.6, 0.05416)};
        \addplot[red, thick, dashdotted] coordinates {(0,0.04595) (1.6, 0.04595)};
        \addplot[
            mark=*, thick, solid, myempirical,
            mark size=1.5pt,
            mark indices={1,3,5,7,9,11,13,15,17}
        ] coordinates {
            (0.0, 0.04672) (0.1, 0.04573) (0.2, 0.04482) (0.3, 0.04401)
            (0.4, 0.04408) (0.5, 0.04368) (0.6, 0.04296) (0.7, 0.04175)
            (0.8, 0.04103) (0.9, 0.04073) (1.0, 0.04046) (1.1, 0.03978)
            (1.2, 0.04002) (1.3, 0.04021) (1.4, 0.03979) (1.5, 0.04035)
            (1.6, 0.0401)
        };
        \end{axis}
        \end{tikzpicture}
    \end{subfigure}

    \vspace{0.1cm} 
    
    \begin{tikzpicture}[font=\footnotesize]
        \coordinate (L1) at (0,0);
        \coordinate (L2) at (2.0,0);
        \coordinate (L3) at (4.0,0);
        
        \draw[blue, thick, dashed] (L1) -- ++(0.5,0) node[right, black] {AE};
        
        \draw[red, thick, dashdotted] (L2) -- ++(0.5,0) node[right, black] {HE};
        
        \draw[thick, solid, myempirical] (L3) -- ++(0.5,0);
        \fill[myempirical] (L3)++(0.25,0) circle (2pt) node[right=0.35cm, black] {Empirical};
    \end{tikzpicture}
    
    \vspace{-0.1cm}

    \caption{\textbf{Performance of the empirical selector across selection thresholds.} As $c$ increases, the selector chooses HE more often. Both mean regret and the error rate increase, while the cost of incorrect selections first decreases and then levels off.}
    \label{fig:selection_threshold}
\end{figure}

Figure~\ref{fig:signal_by_decile} of the main text puts this threshold choice in context. When the SNR is high, AE tends to be as good as or better than HE, so a lower threshold---which makes the selector behave more like AE---would perform better in that regime; when the SNR is low, the reverse holds. But this is not actionable advice: the rule exists precisely because the SNR is unknown in practice. The threshold is therefore best understood as a compromise that makes the selection robust across regimes: at $c=1$, the empirical selector slightly trails AE in the highest-SNR deciles---the price of conservatism---while outperforming it where the signal is weak, tracking the better method across the full range without knowing which regime it is in.

\subsection{Implementation of the Heterogeneity Test}\label{app:sel_test}

The selection procedure uses the heterogeneity test of \cite{imai2025statistical} as a screening step: datasets in which neither tuned estimator detects heterogeneity are excluded, since the AE--HE comparison is moot when the signal is too weak for either method to model heterogeneity reliably.

Our implementation follows the cross-fitting generalization of the test developed by \cite{imai2025statistical}. We create a 5-fold partition of the training sample---distinct from the partition used for hyperparameter tuning---refit each forest on four folds to generate predictions for the held-out fold, and conduct the test on the resulting out-of-fold predictions. No observation is therefore predicted by a model that was trained on it.

One deviation from a fully sample-split protocol is that the observations used for testing were previously used, under a different fold partition, to tune the \texttt{min\_samples\_leaf} hyperparameter. This is a deliberate choice, for two reasons. First, the test serves as a screening device within a model-selection pipeline whose end-to-end performance is what we evaluate, not as a confirmatory inference procedure; if the goal were formal hypothesis testing about heterogeneity, the sample-splitting protocol should be followed as prescribed. Second, reserving a dedicated holdout untouched by tuning would substantially reduce the data available for tuning and for the AE--HE comparison, weakening both estimators in a way that runs counter to our goal of mirroring applied workflows.

The resulting leakage channel is narrow---a single scalar hyperparameter selected from a coarse grid---and it affects AE and HE symmetrically, since both are tuned on the same data and tested under the same partition. As an empirical check, we applied the full procedure to the 200 ACIC datasets with no true effect heterogeneity: the false positive rate was 1.5\% for both AE and HE, below the nominal 5\% level, indicating that the reuse does not make the test anti-conservative in practice.

\section{Additional Data Requirements}
This appendix describes the procedure used to calculate the additional data requirements reported in Figure~\ref{fig:data_efficiency}.

For each dataset, we compare the test MSEs of AE and HE, both trained on the full sample (\(N=4{,}000\)). If the difference is not significant according to a two-sided test at the 5\% level, both models are assigned an effective sample size of 4,000, and the additional data requirement is zero.

If the difference is significant, we determine the minimum training sample size at which the superior model outperforms the inferior model trained on all 4,000 observations. Holding the inferior model's MSE fixed, we use a binary search over the superior model's training sample size. We initialize the search bounds at $N_{\mathrm{low}}=0$ and $N_{\mathrm{high}}=4{,}000$. At each step, we evaluate the candidate size given by the midpoint of the current bounds: we tune and retrain the superior model at this size and evaluate its test MSE. If this MSE is lower than the full-sample inferior model's MSE and the difference is significant, we lower the upper bound to the candidate; otherwise, we raise the lower bound to the candidate. We stop once the bounds are within 250 observations.

The superior model's effective sample size is the final upper bound---the smallest sample size found to yield a statistically significant improvement, up to a resolution of 250 observations. The inferior model's effective sample size remains fixed at 4,000. Denoting the effective sample sizes of HE and AE by \(N^{H}\) and \(N^{A}\), respectively, we calculate the additional-data measure for HE as $(N^{H}-N^{A})/{4{,}000}$. Positive values indicate that HE requires more data than AE, whereas negative values indicate that AE requires more data than HE.

Figure~\ref{fig:data_efficiency} reports the average additional-data measure across datasets within each SNR decile.

\section{Bias--Variance Estimation}
\label{app:bias_variance}

This appendix describes how we estimate the bias--variance decomposition and its subcomponents.

\subsection{Forest-level Decomposition}

The main text presents the bias--variance decomposition at the tree level. For estimation, it is more convenient to rewrite the same decomposition at the forest level. Fix a test point $\mathbf{x}$. Equation~\eqref{eq:tree-variance-decomp} writes the prediction from tree $b$ as
\begin{equation*}
\hat\beta_b^{\mathrm{tree}}
=
\tau_b+\varepsilon_b,
\end{equation*}
where $\tau_b$ is the approximation target of the selected leaf containing $x$, and $\varepsilon_b$ is the estimation error.

Conditional on the training sample $\mathcal S$, we obtain the forest-level representation by averaging this tree-level decomposition over trees,
\begin{equation}
\bar\tau(\mathcal S)
=
\mathbb E_b[\tau_b\mid \mathcal S],
\qquad
\bar\varepsilon(\mathcal S)
=
\mathbb E_b[\varepsilon_b\mid \mathcal S].
\end{equation}
Then the forest prediction is
\begin{equation}
\hat\beta(\mathcal S)
=
\mathbb E_b[\hat\beta_b^{\mathrm{tree}}\mid \mathcal S]
=
\bar\tau(\mathcal S)+\bar\varepsilon(\mathcal S).
\end{equation}

Taking expectation over training samples gives the forest-level bias decomposition:
\begin{equation}
\mathbb E_{\mathcal S}[\hat\beta(\mathcal S)]-\beta
=
\underbrace{
\mathbb E_{\mathcal S}[\bar\tau(\mathcal S)]-\beta
}_{\text{approximation bias}}
+
\underbrace{
\mathbb E_{\mathcal S}[\bar\varepsilon(\mathcal S)]
}_{\text{estimation bias}}.
\end{equation}
This is the forest-level analogue of the tree-level decomposition in the main text, Equation~\eqref{eq:error_decomposition}.

The variance decomposition follows the same representation. For two distinct trees $b\neq b'$, the covariance decomposition in Equation~\eqref{eq:covariance-decomposition} is:
\begin{equation*}
\operatorname{Cov}
\left(
\hat\beta_b^{\mathrm{tree}},
\hat\beta_{b'}^{\mathrm{tree}}
\right)
=
\operatorname{Cov}(\tau_b,\tau_{b'})
+
\operatorname{Cov}(\varepsilon_b,\varepsilon_{b'}) \\
+
\operatorname{Cov}(\tau_b,\varepsilon_{b'})
+
\operatorname{Cov}(\varepsilon_b,\tau_{b'}).
\end{equation*}

We illustrate the argument using the target coupling term. By the law of total covariance,
\begin{equation}
\operatorname{Cov}(\tau_b,\tau_{b'})
=
\operatorname{Cov}_{\mathcal S}
\left(
\mathbb E_b[\tau_b\mid \mathcal S],
\mathbb E_{b'}[\tau_{b'}\mid \mathcal S]
\right) 
+
\mathbb E_{\mathcal S}
\left[
\operatorname{Cov}(\tau_b,\tau_{b'}\mid \mathcal S)
\right].
\end{equation}
Conditional on $\mathcal S$, the two trees are independently generated, so
\begin{equation}
\operatorname{Cov}(\tau_b,\tau_{b'}\mid \mathcal S)=0.
\end{equation}
Moreover,
\begin{equation}
\mathbb E_b[\tau_b\mid \mathcal S]
=
\mathbb E_{b'}[\tau_{b'}\mid \mathcal S]
=
\bar\tau(\mathcal S).
\end{equation}
Therefore,
\begin{equation}
\operatorname{Cov}(\tau_b,\tau_{b'})
=
\operatorname{Var}_{\mathcal S}
\left(
\bar\tau(\mathcal S)
\right).
\end{equation}

Applying the same argument to the remaining terms gives
\begin{equation}
\operatorname{Cov}(\varepsilon_b,\varepsilon_{b'})
=
\operatorname{Var}_{\mathcal S}
\left(
\bar\varepsilon(\mathcal S)
\right),
\end{equation}
and, by symmetry,
\begin{equation}
\operatorname{Cov}(\tau_b,\varepsilon_{b'})
=
\operatorname{Cov}(\varepsilon_b,\tau_{b'})
=
\operatorname{Cov}_{\mathcal S}
\left(
\bar\tau(\mathcal S),
\bar\varepsilon(\mathcal S)
\right).
\end{equation}

Combining these terms yields
\begin{equation}
\operatorname{Cov}
\left(
\hat\beta_b^{\mathrm{tree}},
\hat\beta_{b'}^{\mathrm{tree}}
\right)
=
\underbrace{
\operatorname{Var}_{\mathcal S}
\left(
\bar\tau(\mathcal S)
\right)
}_{\text{target coupling}} 
+
\underbrace{
\operatorname{Var}_{\mathcal S}
\left(
\bar\varepsilon(\mathcal S)
\right)
}_{\text{noise overlap}} \\
+
\underbrace{
2\operatorname{Cov}_{\mathcal S}
\left(
\bar\tau(\mathcal S),
\bar\varepsilon(\mathcal S)
\right)
}_{\text{target--estimation spillover}}.
\end{equation}
Thus, estimating the variance subcomponents reduces to estimating the forest-level approximation target $\bar\tau(\mathcal S)$ and estimation error $\bar\varepsilon(\mathcal S)$ across repeated training samples.

\subsection{Bias and Variance Estimates}

This section describes the estimation procedure for the bias--variance decomposition and its subcomponents.

For each replication $r=1,\ldots,R$, we draw an independent treatment assignment with treatment probability $0.5$ and an independent train/test split of size $4000:802$. We fit the causal forest on the training sample and generate predictions for each test observation.

After fitting the forest, we freeze the learned tree structure. For a test point $\mathbf{x}_i$, let $\hat\ell_b(\mathbf{x}_i)$ denote the leaf containing $\mathbf{x}_i$ in tree $b$. We then use the other test observations as an auxiliary sample, excluding unit $i$ itself. Within each tree, we average the true CATEs of auxiliary observations that fall in the same learned leaf as $\mathbf{x}_i$. The forest-level approximation target is the average of these leaf-level targets across trees:
\begin{equation}
\widehat{\bar\tau}_i^{(r)}
=
\frac{1}{B}
\sum_{b=1}^B
\frac{
\sum_{j\in\mathcal S_{\mathrm{te}}^{(r)},\,j\neq i}
\mathbf 1\{\mathbf{X}_j\in\hat\ell_b(\mathbf{x}_i)\}\beta_j
}{
\sum_{j\in\mathcal S_{\mathrm{te}}^{(r)},\,j\neq i}
\mathbf 1\{\mathbf{X}_j\in\hat\ell_b(\mathbf{x}_i)\}
}.
\end{equation}

The estimation error is the difference between the forest prediction and the estimated approximation target:
\begin{equation}
\widehat{\bar\varepsilon}_i^{(r)}
=
\hat\beta_i^{(r)}
-
\widehat{\bar\tau}_i^{(r)}.
\end{equation}

We repeat this procedure $R=600$ times. Each unit appears in the test set approximately 100 times on average. Let $\mathcal R_i$ denote the set of replications in which unit $i$ appears in the test set, and let $R_i=|\mathcal R_i|$. All pointwise quantities are computed using only replications in $\mathcal R_i$. 

Define the empirical means
\begin{equation}
\bar\tau_i
=
\mathbb E_{\mathcal R_i}
\left[
\widehat{\bar\tau}_i^{(r)}
\right],
\qquad
\bar\varepsilon_i
=
\mathbb E_{\mathcal R_i}
\left[
\widehat{\bar\varepsilon}_i^{(r)}
\right],
\qquad
\bar\beta_i
=
\mathbb E_{\mathcal R_i}
\left[
\hat\beta_i^{(r)}
\right].
\end{equation}

The squared bias is estimated as
\begin{equation}
\widehat{\operatorname{Bias}}_i^2
=
(\bar\beta_i-\beta_i)^2.
\end{equation}
It decomposes into squared approximation bias, squared estimation bias, and their interaction:
\begin{equation}
\widehat{\operatorname{Bias}}_i^2
=
\underbrace{
(\bar\tau_i-\beta_i)^2
}_{\text{squared approximation bias}}
+
\underbrace{
\bar\varepsilon_i^2
}_{\text{squared estimation bias}}
+
\underbrace{
2(\bar\tau_i-\beta_i)\bar\varepsilon_i
}_{\text{bias interaction}}.
\end{equation}

The variance is estimated as
\begin{equation}
\widehat{\operatorname{Var}}_i
=
\widehat{\operatorname{Var}}_{\mathcal R_i}
\left(
\hat\beta_i^{(r)}
\right).
\end{equation}
It decomposes into target coupling, noise overlap, and target--estimation spillover:
\begin{equation}
\widehat{\operatorname{Var}}_i
=
\underbrace{
\widehat{\operatorname{Var}}_{\mathcal R_i}
\left(
\widehat{\bar\tau}_i^{(r)}
\right)
}_{\text{target coupling}}
+
\underbrace{
\widehat{\operatorname{Var}}_{\mathcal R_i}
\left(
\widehat{\bar\varepsilon}_i^{(r)}
\right)
}_{\text{noise overlap}}
+
\underbrace{
2\,\widehat{\operatorname{Cov}}_{\mathcal R_i}
\left(
\widehat{\bar\tau}_i^{(r)},
\widehat{\bar\varepsilon}_i^{(r)}
\right)
}_{\text{target--estimation spillover}}.
\end{equation}
Here, $\mathbb E_{\mathcal R_i}$, $\widehat{\operatorname{Var}}_{\mathcal R_i}$, and $\widehat{\operatorname{Cov}}_{\mathcal R_i}$ denote the empirical mean, variance, and covariance over replications $r\in\mathcal R_i$.

Together, these estimates provide a pointwise decomposition of prediction error. In the main analysis, we average the pointwise components across units and normalize them by the variance of the CATE over the full dataset. 

One caveat applies to the variance subcomponents. Because $\widehat{\bar\tau}_i^{(r)}$ is computed from a finite auxiliary sample, it carries sampling noise of its own. Writing $\widehat{\bar\tau}_i^{(r)} = \bar\tau_i^{(r)} + u_i^{(r)}$, we also have that $\widehat{\bar\varepsilon}_i^{(r)} = \bar\varepsilon_i^{(r)} - u_i^{(r)}$. If $u$ is uncorrelated with $\bar\tau$ and $\bar\varepsilon$, the estimated target coupling and noise overlap are each inflated by $\mathrm{Var}(u)$ and the estimated spillover is displaced downward by $2\,\mathrm{Var}(u)$; the estimated total variance is unaffected. The levels of the individual subcomponents should therefore not be read as estimates of the underlying population quantities. This applies in particular to the target--estimation spillover, which is displaced toward negative values.

Comparisons between AE and HE are on firmer ground. Because $\mathrm{Var}(u)$ decreases in the number of auxiliary observations per leaf, it is larger for whichever estimator produces the finer partition---in the self-optimal regime, HE. The displacement therefore works against the reported differences in target coupling and noise overlap, both of which are estimated to be lower for HE despite HE carrying the larger upward displacement, so those comparisons are conservative. Comparisons across components require more care. Target coupling and noise overlap are inflated by the same amount, so their difference is unaffected and the finding that noise overlap dominates the AE--HE variance gap does not depend on the displacement.

\section{Honest vs. Adaptive Lasso on the ACIC Datasets}\label{app:lasso}

Honesty is a design principle that is not exclusive to causal forest. It applies to any learning procedure that separates variable selection (or space partitioning) from parameter estimation. Lasso illustrates this idea well, as it combines variable selection with coefficient estimation. When both tasks are performed on the same sample, we obtain an adaptive Lasso, in which selection errors propagate into the estimated coefficients. In contrast, an honest Lasso decouples the two stages by using separate samples for selection and estimation, thereby mitigating overfitting.

This principle also extends to meta-learners in causal inference, whose behavior inherits the properties of their base learners. For example, a T-learner built with random forests is honest when honest forests are used as base learners and adaptive otherwise. We use Lasso as a representative example to compare adaptive and honest implementations on the ACIC datasets.

We consider the following linear interaction model~\citep{imbens2015causal}
\begin{equation}
Y_i
= \alpha_0 + \gamma_0 T_i + \mathbf{X}_i^\top \boldsymbol{\alpha}
+ T_i(\mathbf{X}_i-\bar{\mathbf{X}})^\top \boldsymbol{\gamma}
+ \epsilon_i ,
\end{equation}
where $\alpha_0$, $\gamma_0$, $\boldsymbol{\alpha}$, and $\boldsymbol{\gamma}$ are coefficients to be learned, and $\bar{\mathbf{X}}$ denotes the vector of covariate means. This model yields the predicted individual treatment effect
\begin{equation}
\hat{\beta}_i
= \hat{\gamma}_0
+ (\mathbf{X}_i-\bar{\mathbf{X}})^\top \hat{\boldsymbol{\gamma}} .
\end{equation}

For adaptive Lasso, we fit a single Lasso model on the full training sample, performing variable selection and coefficient estimation simultaneously within one regression.

For honest Lasso, we separate selection and estimation using disjoint samples and implement this via 2-fold cross-fitting. The training data are randomly partitioned into two equal folds, $\mathcal{S}_{\text{sel}}$ and $\mathcal{S}_{\text{es}}$.
\begin{enumerate}
    \item \textbf{Selection:} Fit Lasso on $\mathcal{S}_{\text{sel}}$ to identify the active set of variables.
    \item \textbf{Estimation:} Fit ordinary least squares on $\mathcal{S}_{\text{es}}$ using only the selected variables to estimate coefficients.
\end{enumerate}
We then swap the roles of the two folds and average predictions across the two fits. This cross-fitting scheme is essentially an ensemble of honest estimators, which recovers much of the efficiency lost in a single honest fit by ensuring that every observation contributes to both stages. For comparison, we also evaluate a single-fit honest Lasso, which does not swap the folds and uses only half of the data for each task.

For both methods, the Lasso $L_1$ regularization parameter ($\lambda$) is tuned over a log-spaced grid of 30 values from $10^{-3}$ to approximately $3$, with the optimal value selected by minimizing the transformed MSE. For honest Lasso, $\lambda$ is tuned separately for the two selection models.

\begin{figure}[t]
    \centering
    \begin{tikzpicture}
        \begin{axis}[
            width=12cm, height=8cm,
            xlabel={SNR Decile},
            ylabel={Proportion of Signal ($S^2$)},
            ymin=0.35, ymax=0.62,
            xtick={0,...,9},
            xticklabels={1, 2, 3, 4, 5, 6, 7, 8, 9, 10},
            grid=major,
            grid style={dashed, gray!30},
            legend pos=south east,
            legend cell align=left,
            legend style={font=\small}, 
        ]
        \addplot+[mark=*, thick, solid, mark size=2pt] coordinates {
            (0, 0.41762)
            (1, 0.52733)
            (2, 0.53172)
            (3, 0.54545)
            (4, 0.54857)
            (5, 0.56257)
            (6, 0.54765)
            (7, 0.55956)
            (8, 0.57084)
            (9, 0.59603)
        };
        \addlegendentry{AE}
        \addplot+[mark=square*, thick, dashed, mark size=2pt] coordinates {
            (0, 0.44449)
            (1, 0.51706)
            (2, 0.52547)
            (3, 0.54239)
            (4, 0.54332)
            (5, 0.55431)
            (6, 0.54463)
            (7, 0.5558)
            (8, 0.56921)
            (9, 0.59128)
        };
        \addlegendentry{HE (cross-fit)}
        \addplot+[myempirical, mark=triangle*, thick, dashdotted, 
        mark size=2.5pt, mark options={solid}] coordinates {
            (0, 0.37988)
            (1, 0.47688)
            (2, 0.49475)
            (3, 0.5179)
            (4, 0.52192)
            (5, 0.53741)
            (6, 0.52882)
            (7, 0.54329)
            (8, 0.55709)
            (9, 0.5809)
        };
        \addlegendentry{HE (single-fit)}
        \end{axis}
    \end{tikzpicture}
    \caption{\textbf{Proportion of explained variance ($S^2$) across SNR deciles for adaptive, cross-fit honest, and single-fit honest Lasso models.} AE underperforms HE in the lowest SNR decile and is preferable when SNR is higher. This highlights the regularization effect of honesty. The weak performance of the single-fit HE emphasizes the efficiency loss of sample splitting, which is mitigated via cross-fitting.}
    \label{fig:acic-lasso}
\end{figure}

Figure~\ref{fig:acic-lasso} reports the average $S^2$ across SNR deciles for adaptive Lasso, honest Lasso with cross-fit, and honest Lasso with single-fit. The adaptive and cross-fitted honest versions exhibit a pattern similar to causal forests: honesty performs better in the lowest-SNR regime, whereas adaptivity becomes advantageous as the SNR increases. This pattern reinforces our core insight that honesty acts as a form of regularization.

The performance gap is smaller than for forests because Lasso combines a linear structure with $L_1$ shrinkage. This constraint already limits flexibility and overfitting, so the additional regularization provided by honesty has a more modest effect. Moreover, even half the sample is often enough to identify relevant predictors, which reduces the cost of sample splitting.

Finally, the single-split honest Lasso illustrates the cost of honesty: using only half the data at each stage increases variance. Cross-fitting mitigates this loss by bagging two honest estimators, allowing all observations to contribute. In this sense, moving from a single-fit to a cross-fit honest Lasso is analogous to moving from a causal tree to a forest---aggregation offsets the variance introduced by sample splitting while preserving the regularization benefit. A key practical takeaway is that, whenever honest estimation is used, it should be paired with bagging.

\section{Learning Curve Analysis for the MegaFon Dataset}\label{app:megafon}

To complement the smaller sample sizes in the ACIC datasets, we extend our comparison of honest and adaptive forests to a large scale setting. We use the MegaFon dataset from the 2021 MegaFon Uplift Competition, a synthetic dataset designed to mimic realistic telecom customer behavior.\footnote{MegaFon dataset URL: \url{https://ods.ai/tracks/df21-megafon/competitions/megafon-df21-comp/data}} It contains 600,000 observations and 50 features, which allows us to evaluate model performance across a wide range of training sample sizes. The treatment indicates whether a customer receives an offer (treatment rate 50\%), and the outcome records whether the desired action is taken. The average treatment effect is 4.95\%.

Our goal is to assess how adaptive and honest causal forests scale with training data size. In the ACIC datasets, different datasets differ in signal strength and therefore the SNR. In the MegaFon setting, we instead vary the SNR through the training sample size: less data leads to more sampling error---and thus a lower SNR from the learner's perspective---while larger training sets increase the SNR. This setup lets us study performance along a learning curve.

We use 5×5 nested cross-validation to obtain out-of-sample estimates for every observation.
\begin{itemize}
    \item \textbf{Outer loop (evaluation)}: We split the data into five outer folds. In each iteration, one fold is held out as a test set, and the remaining four folds form the training pool. From this pool, we draw training subsets of increasing sizes
    \[
        N \in \{1\mathrm{K}, 2\mathrm{K}, 5\mathrm{K}, 10\mathrm{K}, 20\mathrm{K}, 50\mathrm{K}, 100\mathrm{K}, 200\mathrm{K}, 480\mathrm{K}\}.
    \]
    For each $N$, we train a model on the subset and evaluate it on the held-out outer test fold. This procedure measures how predictive performance scales with data availability.
    \item \textbf{Inner loop (hyperparameter tuning)}: For each training subset of size $N$, we perform 5-fold inner cross-validation to tune the \texttt{min\_samples\_leaf} hyperparameter over $\{10, 20, 50, 100, 200\}$. We select the value that minimizes the MSE on transformed outcomes within the validation set, as discussed in Section~\ref{sec:design}. The model was then retrained on the full subset using the optimal hyperparameter before final evaluation on the outer test fold.
\end{itemize}

Because the true CATE is unknown in the MegaFon dataset, we evaluate performance using the MSE of transformed outcomes ($Z$) on the full dataset. We adapt the $S^2$ metric as
\begin{equation}
S^2_z = 1 - \frac{\mathrm{MSE}}{\mathrm{Var}(Z)},
\end{equation}
where $\mathrm{Var}(Z)$ is the variance of the transformed outcomes across the full dataset. In this context, $S^2_z$ measures the proportion of variance in the transformed outcomes explained by the model.

\begin{figure}[t]
    \centering
    \pgfplotsset{
        learningcurvestyle/.style={
            width=\linewidth, 
            height=6.4cm,  
            xlabel={Training data size (log scale)}, 
            xmin=900, xmax=550000,
            xtick={1000, 2000, 5000, 10000, 20000, 50000, 100000, 200000, 480000},
            xticklabels={1K, 2K, 5K, 10K, 20K, 50K, 100K, 200K, 480K},
            tick label style={font=\scriptsize},
            legend style={font=\footnotesize}, 
            grid=major,
            grid style={dashed, gray!30},
        }
    }
    \begin{subfigure}[b]{0.495\textwidth}
        \caption{$S^2_z$ Score}\label{fig:megafon-s-squared}
        \begin{tikzpicture}
            \begin{semilogxaxis}[
                learningcurvestyle, 
                ytick={0, 1, 2, 3},
                yticklabels={0\%, 1\%, 2\%, 3\%},
                legend pos=north west,
            ]
            
            \addplot[thick, solid, color=blue, mark=*, mark size=1.5pt] coordinates {
                (1000, -0.2126933) (2000, -0.2594963) (5000, 0.1271465)
                (10000, 0.8327568) (20000, 1.4082899) (50000, 1.9526917)
                (100000, 2.3074067) (200000, 2.6066197) (480000, 2.9840914)
            };
            \addlegendentry{AE}
            
            \addplot[thick, dashed, color=red, mark=square*, mark size=1.5pt] coordinates {
                (1000, -0.0936131) (2000, -0.0003021) (5000, 0.1061923)
                (10000, 0.3345138) (20000, 0.7919721) (50000, 1.3108292)
                (100000, 1.7150443) (200000, 2.1022205) (480000, 2.5180984)
            };
            \addlegendentry{HE}
            \end{semilogxaxis}
        \end{tikzpicture}
    \end{subfigure}
    \hfill 
    \begin{subfigure}[b]{0.495\textwidth}
        \caption{$S^2_z$ Difference (AE - HE)}\label{fig:megafon-s-squared-diff}
        \begin{tikzpicture}
            \begin{semilogxaxis}[
                learningcurvestyle, 
                ytick={-0.2, 0, 0.2, 0.4, 0.6},
                yticklabels={-0.2\%, 0\%, 0.2\%, 0.4\%, 0.6\%},
            ]
            
            \addplot[dashed, black, thick, forget plot] coordinates {(900, 0) (550000, 0)};
                            
            \addplot[fill=blue, fill opacity=0.2, draw=none, forget plot] coordinates {
                (1000, -0.1372386) (2000, -0.3045526) (5000, -0.0112032)
                (10000, 0.4501323) (20000, 0.5644091) (50000, 0.602908)
                (100000, 0.5535025) (200000, 0.4668595) (480000, 0.425039) 
                (480000, 0.506947) (200000, 0.5419389) (100000, 0.6312222)
                (50000, 0.6808169) (20000, 0.6682264) (10000, 0.5463536)
                (5000, 0.0531118) (2000, -0.2138356) (1000, -0.1009219) 
            };

            \addplot[thick, color=blue, mark=*, mark size=1.5pt] coordinates {
                (1000, -0.1190803) (2000, -0.2591941) (5000, 0.0209543)
                (10000, 0.4982429) (20000, 0.6163178) (50000, 0.6418625)
                (100000, 0.5923623) (200000, 0.5043992) (480000, 0.465993)
            };
            \end{semilogxaxis}
        \end{tikzpicture}
    \end{subfigure}
    \caption{\textbf{Learning curves for causal forests on the MegaFon dataset.} The panels illustrate the out-of-sample $S^2_z$ (left) and the average difference in squared errors with 95\% confidence intervals (right) across training sample sizes ($N$). The results closely mirror the patterns observed on the ACIC datasets when training size is interpreted as a proxy for the SNR: HE holds a slight advantage in the high-noise, small-sample regime ($N \le 2\mathrm{K}$), while AE achieves a significant performance lead as data size increases.}
\end{figure}

Figure~\ref{fig:megafon-s-squared} reports the $S^2_z$ of adaptive and honest forests across training sizes, and Figure~\ref{fig:megafon-s-squared-diff} shows the average difference in squared errors, alongside 95\% confidence intervals. Note that $S^2_z$ values remain small due to the high variance in transformed outcomes.

The results closely mirror the patterns observed on the ACIC datasets when training size is interpreted as a proxy for the SNR. In the small-sample regime ($N \le 2\mathrm{K}$), HE performs better than AE, although the negative $S^2_z$ values indicate that both methods overfit the noise and underperform an ATE baseline. As $N$ grows, AE overtakes HE, expanding its lead as the SNR improves. Finally, at large sample sizes ($N \ge 50\mathrm{K}$), this performance gap gradually narrows.

These results show that AE's advantage is not a small-sample artifact; it persists in large samples. But sample size is only one determinant of the SNR, which also depends on the data-generating process. Thus, the performance gap between AE and HE reflects differences in underlying SNR rather than sample size per se. The MegaFon learning curves confirm that AE tends to outperform in higher-SNR regimes, while HE works better when the signal is weak.

\section{Literature Review}
\label{app:lit_review}

This appendix reviews work on causal trees and causal forests, with a focus on honest estimation.

\cite{athey2016recursive} adapt the CART algorithm to the problem of estimating heterogeneous treatment effects. Their key contributions are a splitting criterion designed to detect effect heterogeneity, and the introduction of honest estimation, in which separate samples are used to construct the tree partition and to estimate effects within leaves. The motivation for honesty is to reduce bias from adaptive overfitting and to facilitate valid statistical inference after partition selection.

\citet{wager2018estimation} extend this framework to random forests, showing that ensembles of honest trees yield consistent effect estimates and enable asymptotically valid inference. This approach is further generalized in the generalized random forest framework~\citep{athey2019generalized}, which casts forest methods as adaptive weighting schemes for a broad class of local moment conditions, allowing effect estimation to be embedded in a more flexible and unified framework.

Recent theoretical work helps explain why honesty is appealing, but also why it is not a safeguard against poor CATE estimates. When the data used to choose splits is separated from the data used to estimate outcomes within leaves, trees can achieve consistent estimation under standard regularity conditions. Specifically, ~\cite{bladt2026consistency} show that honest regression trees and random forests can consistently estimate conditional expectations when leaves become more localized in feature space---reducing approximation error---while still containing enough observations for stable within-leaf estimation. 

However, this consistency does not guarantee good finite-sample performance, particularly when the resulting leaves are unstable or poorly aligned with the underlying heterogeneity. For instance,~\cite{cattaneo2022pointwise} show that CART-style greedy splitting has a persistent ``end-cut preference,'' producing imbalanced leaves with non-vanishing probability even when pruning is allowed; ensembling via random forests can mitigate the impact of this issue on predictions, but does not eliminate it.~\cite{cattaneo2025honest} extend this analysis to derive accuracy limits for widely used causal-tree estimators and splitting rules, showing that honest estimation does not address these small-leaf pathologies. Taken together, these results suggest that honesty can simplify theoretical analysis and enable valid inference, but the resulting partitions may still provide poor approximations for individual CATEs.

Empirical work suggests that whether honest or adaptive estimation works better depends on the data-generating process and on how model complexity (such as tree depth) is tuned. \cite{havelka2022honesty} compares the two approaches in simulations and finds that honesty tends to perform better in settings where subgroup selection is prone to overfitting---for example, when treatment effect heterogeneity is relatively mild---while adaptive estimation can perform better when effects are sharply concentrated in specific subgroups. Although not framed in these terms by Havelka, this pattern is consistent with interpreting honesty as a form of regularization, reducing variance from overfitting at the cost of potentially higher approximation bias.

Similarly,~\cite{prodan2025analyzing} shows that whether honesty improves accuracy depends on how tree depth and leaf size are tuned, as these hyperparameters shape the bias–variance trade-off. In noisy or high-dimensional settings, pairing honesty with shallower trees and larger leaves improves performance by controlling variance. But when treatment effects are highly nonlinear and involve many interactions, deeper trees and more adaptive (non-honest) approaches can reduce approximation bias and achieve lower overall error. These findings suggest that the empirical question is not whether honest or adaptive estimation is universally preferable, but under which regimes---defined jointly by noise, the nature of effect heterogeneity, sample size, and tuning choices---one approach is better than the other.

Beyond treatment effect estimation, related work also examines honesty in forest-based decision making, including stochastic optimization and policy learning. In this literature, sample splitting between tree construction and decision estimation is often imposed to make asymptotic optimality guarantees tractable. \citet{kallus2023stochastic} note that this requirement is largely technical---introduced to simplify theoretical analysis---and can reduce performance in finite samples, a pattern they corroborate with empirical comparisons.

Taken together, these findings---including ours---imply that the impact of honest estimation on accuracy depends on context and should be evaluated within each specific application. Yet in practice, it is often treated as a default rather than a deliberate choice. In our review of 45 studies published in INFORMS journals (Appendix~\ref{app:causal_forest_applications}), fewer than half explicitly reported using honest estimation (22 studies), fewer than a third justified that choice (14 studies), and only two studies considered potential accuracy trade-offs or compared it to adaptive estimation~\citep{fernandez2025observational,kallus2023stochastic}. In both cases, honest estimation often reduced performance. Among the 14 studies that provided a justification, most focused on avoiding overfitting or ensuring consistency, rather than on whether honesty improved performance in the specific application. This pattern suggests that honest estimation is frequently adopted without careful evaluation. We hope our study encourages researchers to treat honesty as a design choice---one that merits explicit reasoning and empirical assessment. 

\section{Applications of Causal Tree and Forest Methods}\label{app:causal_forest_applications}

Table~\ref{table:causal_forest_applications} presents 45 papers published in INFORMS journals (2019\textendash2025) that use causal trees or causal forests. The majority employ these methods to estimate individual-level effects, which are then used for other analyses.

{\footnotesize
\setlength{\arrayrulewidth}{0.3pt}
\setlength{\tabcolsep}{4pt}

\begin{longtable}{P{0.57\linewidth} P{0.18\linewidth} P{0.18\linewidth}}
\caption{Research Papers Using Causal Tree/Forest Methods}
\label{table:causal_forest_applications}\\
\toprule
\textbf{Paper} & \textbf{Journal} & \textbf{Application} \\
\midrule
\endfirsthead

\toprule
\textbf{Paper} & \textbf{Journal} & \textbf{Application} \\
\midrule
\endhead

\midrule
\endfoot

\bottomrule
\endlastfoot

Bertsimas et al. (2019): Optimal Prescriptive Trees & INFORMS Journal on Optimization & Treatment assignment \\ \midrule
Luo et al. (2019): When and How to Leverage E-commerce Cart Targeting: The Relative and Moderated Effects of Scarcity and Price Incentives with a Two-Stage Field Experiment and Causal Forest Optimization & Information Systems Research & Effect heterogeneity analysis \\ \midrule
Ge et al. (2021): Human–Robot Interaction: When Investors Adjust the Usage of Robo-Advisors in Peer-to-Peer Lending & Information Systems Research & ATE estimation \\ \midrule
Hermosilla (2021): Rushed Innovation: Evidence from Drug Licensing & Management Science & Effect heterogeneity analysis \\ \midrule
Kallus and Zhou (2021): Minimax-Optimal Policy Learning Under Unobserved Confounding & Management Science & Treatment assignment \\ \midrule
Fernández-Loría and Provost (2022): Causal Decision Making and Causal Effect Estimation Are Not the Same...and Why It Matters & INFORMS Journal on Data Science & Treatment assignment \\ \midrule
Choudhary et al. (2022): Nudging Drivers to Safety: Evidence from a Field Experiment & Management Science & Effect heterogeneity analysis \\ \midrule
Cui and Davis (2022): Tax-Induced Inequalities in the Sharing Economy & Management Science & Effect heterogeneity analysis \\ \midrule
Godinho De Matos and Adjerid (2022): Consumer Consent and Firm Targeting After GDPR: The Case of a Large Telecom Provider & Management Science & Effect heterogeneity analysis \\ \midrule
Pattabhiramaiah et al. (2022): Spillovers from Online Engagement: How a Newspaper Subscriber’s Activation of Digital Paywall Access Affects Her Retention and Subscription Revenue & Management Science & Effect heterogeneity analysis \\ \midrule
Wang (2022): The Effect of Medicaid Expansion on Wait Time in the Emergency Department & Management Science & Causal forest-based method development \\ \midrule
Wang et al. (2022): An Instrumental Variable Forest Approach for Detecting Heterogeneous Treatment Effects in Observational Studies & Management Science & Causal tree-based method development \\ \midrule
Fernández-Loría et al. (2023): A Comparison of Methods for Treatment Assignment with an Application to Playlist Generation & Information Systems Research & Treatment assignment \\ \midrule
Cohen et al. (2023): Managing Airfares Under Competition: Insights from a Field Experiment & Management Science & Effect heterogeneity analysis \\ \midrule
Kallus and Mao (2023): Stochastic Optimization Forests & Management Science & Causal forest-based method development \\ \midrule
Serra-Garcia and Szech (2023): Incentives and Defaults Can Increase COVID-19 Vaccine Intentions and Test Demand & Management Science & Effect heterogeneity analysis \\ \midrule
Unal and Park (2023): Fewer Clicks, More Purchases & Management Science & Effect heterogeneity analysis \\ \midrule
Yoganarasimhan et al. (2023): Design and Evaluation of Optimal Free Trials & Management Science & Treatment assignment \\ \midrule
Zhang and Luo (2023): Can Consumer-Posted Photos Serve as a Leading Indicator of Restaurant Survival? Evidence from Yelp & Management Science & ATE estimation \\ \midrule
Zhang et al. (2023): Uncovering Synergy and Dysergy in Consumer Reviews: A Machine Learning Approach & Management Science & Effect heterogeneity analysis \\ \midrule
Borenstein et al. (2023): Ancillary Services in Targeted Advertising: From Prediction to Prescription & Manufacturing \& Service Operations Management & Effect heterogeneity analysis \\ \midrule
Wang et al. (2023): Personalized Healthcare Outcome Analysis of Cardiovascular Surgical Procedures & Manufacturing \& Service Operations Management & Causal tree-based method development \\ \midrule
Zhou et al. (2023): Offline Multi-Action Policy Learning: Generalization and Optimization & Operations Research & Treatment assignment \\ \midrule
Ayabakan et al. (2024): Impact of Telehealth and Process Virtualization on Healthcare Utilization & Information Systems Research & ATE estimation \\ \midrule
Leng and Dimmery (2024): Calibration of Heterogeneous Treatment Effects in Randomized Experiments & Information Systems Research & Correction for causal forest effect estimates \\ \midrule
Bundorf et al. (2024): How Do Consumers Interact with Digital Expert Advice? Experimental Evidence from Health Insurance & Management Science & Effect heterogeneity analysis \\ \midrule
Kraus et al. (2024): Data-Driven Allocation of Preventive Care with Application to Diabetes Mellitus Type II & Manufacturing \& Service Operations Management & Treatment prioritization \\ \midrule
Pauphilet (2024): Robust and Heterogeneous Odds Ratio: Estimating Price Sensitivity for Unbought Items & Manufacturing \& Service Operations Management & Effect heterogeneity
analysis \\ \midrule
Berman and Feit (2024): Latent Stratification for Incrementality Experiments & Marketing Science & ATE estimation \\ \midrule
Huang and Ascarza (2024): Doing More with Less: Overcoming Ineffective Long-Term Targeting Using Short-Term Signals & Marketing Science & Treatment prioritization \\ \midrule
Turjeman and Feinberg (2024): When the Data Are Out: Measuring Behavioral Changes Following a Data Breach & Marketing Science & Causal forest-based method development \\ \midrule
Fernández-Loría and Provost (2025): Observational vs. Experimental Data When Making Automated Decisions Using Machine Learning & INFORMS Journal on Data Science & Treatment assignment \\ \midrule
Kim et al. (2025): Working Daily, Paid Monthly? Effects of On-Demand Wage Access on the Financial Engagement of Low-Wage Workers & Information Systems Research & ATE estimation \\ \midrule
Alyakoob and Rahman (2025): Market Design Choices, Racial Discrimination, and Equitable Microentrepreneurship in Digital Marketplaces & Management Science & Effect heterogeneity analysis \\ \midrule
Cao et al. (2025): Antisocial Responses to the ``Coal to Gas'' Regulation: An Unintended Consequence of a Residential Energy Policy & Management Science & Effect heterogeneity analysis \\ \midrule
Gulen et al. (2025): Balancing External vs. Internal Validity: An Application of Causal Forest in Finance & Management Science & Effect heterogeneity analysis \\ \midrule
Opitz et al. (2025): The Algorithmic Assignment of Incentive Schemes & Management Science & Treatment assignment \\ \midrule
Long et al. (2025): The Choice Overload Effect in Online Recommender Systems & Manufacturing \& Service Operations Management & Effect heterogeneity analysis \\ \midrule
Luo et al. (2025): From Stopping to Shopping: A Field Experiment on Free Return and Free Shipping Retargeting Policies in Online Retail Operations & Manufacturing \& Service Operations Management & Effect heterogeneity analysis \\ \midrule
Pourghannad and Wang (2025): Matching Patients with Surgeons: Heterogeneous Effects of Surgical Volume on Surgery Duration & Manufacturing \& Service Operations Management & Effect heterogeneity analysis \\ \midrule
Huang and Morozov (2025): The Promotional Effects of Live Streams by Twitch Influencers & Marketing Science & Effect heterogeneity analysis \\ \midrule
Kaul et al. (2025): Call Me Maybe: Does Customer Feedback Seeking Impact Nonsolicited Customers? & Marketing Science & Effect heterogeneity analysis \\ \midrule
Lin and Strulov-Shlain (2025): Choice Architecture, Privacy Valuations, and Selection Bias in Consumer Data & Marketing Science & Effect heterogeneity analysis \\ \midrule
Rafieian et al. (2025): Multiobjective Personalization of Marketing Interventions & Marketing Science & Effect heterogeneity analysis \\ \midrule
Von Zahn et al. (2025): Smart Green Nudging: Reducing Product Returns Through Digital Footprints and Causal Machine Learning & Marketing Science & Treatment assignment \\ 
\end{longtable}
}

\end{document}